\documentclass[letterpaper, 10 pt, conference]{ieeeconf}
\IEEEoverridecommandlockouts
\usepackage{cite}
\usepackage{amsmath,amssymb,amsfonts}
\usepackage{graphicx}
\usepackage{xcolor}
\usepackage{booktabs}
\usepackage{multirow}
\usepackage{tikz}
\usepackage{url}
\usetikzlibrary{arrows.meta,positioning,calc,shapes.geometric}

\usepackage{pifont}
\newcommand{\cmark}{\ding{51}}

\title{\LARGE\bf
RecMorph: Topology-Guided Spatial Recurrence\\
for Generalized Morphology Control}

\author{
Quanrui Rao$^{1}$,
Yong Liu$^{2}$,
Xueming Xiao$^{3}$,
Yingbo Luo$^{1}$,
Kun Wu$^{2}$,
Zhenyu Xu$^{2}$,
and Meibao Yao$^{1,*}$%
\thanks{$^{1}$Quanrui Rao, Yingbo Luo, and Meibao Yao are with
Jilin University, China.
Emails: \{raoqr25, luoyb23\}@mails.jlu.edu.cn,
meibaoyao@jlu.edu.cn.}%
\thanks{$^{2}$Yong Liu, Kun Wu, and Zhenyu Xu are with the
State Key Laboratory of Special Vehicle Design and Manufacturing
Integration Technology, China.
Emails: liuyong9640@163.com,
wukunhenry@163.com,
76720123@qq.com.}%
\thanks{$^{3}$Xueming Xiao is with
Changchun University of Science and Technology, China.
Email: alexcapshow@gmail.com.}%
\thanks{$^{*}$Corresponding author: Meibao Yao.}%
}

\begin{document}
\maketitle
\thispagestyle{empty}
\pagestyle{empty}

\begin{abstract}
Generalized morphology control requires a single policy to transform
information across limbs with different physical roles, coordinate
whole-body motion, and remain efficient as body size grows. Existing
communication mechanisms address these requirements only partially. We introduce \textbf{RecMorph}, a topology-guided spatial recurrent
architecture that uses recurrent sequence computation to jointly perform
cross-limb communication and representation transformation. A depth-first
traversal converts the kinematic tree into a morphology-derived sequence,
along which shared bidirectional transitions progressively transform limb
information before action decoding. Residual preservation, RMS
normalization, and input-dependent channel modulation stabilize this
repeated spatial transformation, yielding linear token complexity at fixed
model width and depth. Across five UNIMAL tasks, RecMorph achieves the strongest mean final
training performance among the evaluated generalized morphology
controllers and the highest measured inference throughput on FT, while
generalizing to unseen variations and bodies with up to 30 limbs. We
further migrate representative generalized controllers from UNIMAL
benchmarks to a four-platform quadruped setting. RecMorph achieves the
best macro-averaged performance under nominal and high friction, reduces
nominal velocity RMSE by 43.5\% relative to specialist MLPs, and one shared
policy completes 40 physical Go1/Go2 trials without falls. These results
show that topology-guided recurrent transformation provides an effective
and efficient communication mechanism for Generalized Morphology Control
and remains effective when transferred from procedural bodies to physical
robot platforms. Code and experimental resources are publicly available at
\url{https://github.com/quanruirao/RecMorph}.
\end{abstract}

\section{Introduction}

\begin{figure}[t]

    \centering
    \includegraphics[width=\columnwidth]{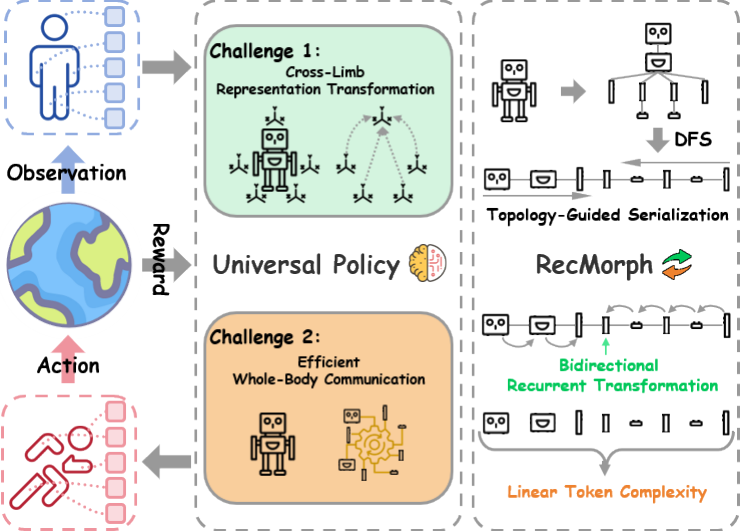}
    \caption{\textbf{Motivation and overview of RecMorph.}
RecMorph uses topology-ordered recurrent transformation to achieve
cross-limb contextualization, whole-body communication, and linear token
complexity.}
    \label{fig:controller}
  \vspace{-0.2in}    
\end{figure}

A robot's morphology determines how local motion contributes to whole-body
behavior. The same joint velocity can propel one body, stabilize another,
or destabilize a third. Learning one controller across these bodies
therefore requires more than accommodating a variable number of actuators:
information expressed in one limb's physical context must be converted
into representations that are useful for controlling other parts of the
body. This creates three coupled requirements for Generalized Morphology Control:
the policy must transform information across heterogeneous limb contexts,
coordinate the complete body, and keep this communication efficient as the
number of limbs grows. Meeting these requirements enables control knowledge to be reused across
robot structures without designing and training an independent controller
for every body.

Existing controllers address different parts of this problem~\cite{wang2018nervenet,huang2020one,gupta2022metamorph,xiong2023universal}.
Graph message passing follows the robot kinematic structure and provides
efficient local communication. Whole-body interaction is obtained by
increasing message-passing depth, so already aggregated neighborhood
representations are repeatedly mixed as the receptive field expands,
which can weaken limb-specific information. Dense self-attention removes
this local receptive-field restriction and gives every limb direct access
to every other limb, but requires quadratic pairwise token interactions.
Moreover, a source limb contributes essentially the same projected value
representation to different targets, with target specificity entering
primarily through attention weights. There is no process of transformation into target-specific limb or movement representations.
Morphology-conditioned projections improve this local representation
mapping, but retain dense attention for body-wide aggregation and its
associated pairwise computation. In this paper, \emph{cross-limb contextualization} denotes constructing a
target-conditioned representation from information originating in
different limb-local physical contexts. The operation may be realized
through direct target-dependent aggregation, morphology-conditioned
transformation, progressive propagation, or a combination of these
mechanisms. We use \emph{cross-limb representation compatibility} to
describe whether the resulting heterogeneous limb representations can
support shared downstream communication and action decoding. These limitations leave a clear need for
a communication operator that performs explicit cross-limb contextualization,
provides whole-body reach, and scales linearly with morphology size.

These requirements motivate recurrent sequence computation. A recurrent
state carries information across an ordered body representation using a
number of transitions that grows linearly with sequence length, while each
transition simultaneously transforms the propagated feature before it
reaches another limb. Communication and representation transformation are
therefore performed by the same operator.
Our proposed \textbf{RecMorph} exploits this property by serializing the kinematic tree with a
depth-first traversal and applying shared bidirectional recurrence along
the resulting morphology-derived sequence. Information originating from a
source limb is progressively transformed before contributing to downstream
limb representations, while the two recurrent directions provide
whole-body communication within each block.
Repeated spatial transformation also introduces a practical challenge:
propagated context must remain stable without overwriting the local state
of the current limb. RecMorph therefore preserves limb-local information
through a residual pathway, controls feature scale with RMSNorm, and uses
input-dependent channel modulation to regulate the propagated
representation before shared action decoding. At fixed width and block
depth, the resulting computation scales linearly with the number of limb
tokens.

Most generalized morphology controllers are developed and compared on
procedurally generated benchmarks such as UNIMAL, where heterogeneous
bodies already share a unified simulation and control interface. It remains
unclear whether architectural advantages observed in this setting persist
when these controllers are transferred to standard robot platforms with
different dynamics and low-level control conventions.
We therefore migrate representative generalized morphology controllers to
a common quadruped benchmark spanning Go1, Go2, ANYmal-B, and ANYmal-C.
Each platform is mapped to the same token-based policy interface, while
its native joint ordering, reference configuration, low-level gains, and
safety limits are preserved in the actuation layer. This setting allows the controller architectures to be compared across
standard robot platforms under a common training protocol. RecMorph retains a substantial advantage over the evaluated generalized
controllers under nominal and high friction, and the shared-policy design
is further deployed on physical Go1 and Go2.

Our contributions are threefold:

(i) We introduce \textbf{RecMorph}, a generalized controller that converts
robot topology into a recurrent communication sequence. Shared
bidirectional transitions perform both cross-limb communication and
representation transformation, providing whole-body interaction with
linear token complexity at fixed network width and depth.

(ii) We develop a stabilized spatial recurrent block that preserves local
limb information during repeated transformation. UNIMAL experiments show
that RecMorph combines strong control performance with high inference
throughput, while controlled studies identify the contributions of
recurrent stabilization and morphology-derived ordering and demonstrate
generalization to unseen and substantially larger bodies.

(iii) We transfer representative Generalized Morphology Control methods
from procedural UNIMAL benchmarks to a common four-quadruped setting
and develop the interface required for shared-policy control across
heterogeneous robot platforms. RecMorph achieves the strongest aggregate
performance under nominal and high friction and is further deployed with
one shared policy on physical Go1 and Go2.

\section{Related Work}

\textbf{Generalized morphology control.}
Graph-based controllers encode robot structure directly in the policy.
NerveNet propagates messages between connected body modules
~\cite{wang2018nervenet}, while shared modular policies perform bottom-up
and top-down communication over the kinematic tree
~\cite{huang2020one}. Token-based controllers instead construct global
interactions among limb representations. MetaMorph applies self-attention
to DFS-ordered limb tokens~\cite{gupta2022metamorph}; SWAT and Body
Transformer introduce additional structural constraints into attention
~\cite{hong2022structure,sferrazza2024body}; and GCNT combines morphology
encoding with graph-based communication~\cite{luo2025gcnt}. ModuMorph
generates morphology-conditioned projections to adapt limb
representations before global aggregation~\cite{xiong2023universal}.
RecMorph introduces a recurrent communication operator in which robot
topology determines the processing order and shared bidirectional
transitions progressively transform information between limb contexts. An operator-level comparison with attention- and graph-based morphology controllers is provided in Appendix~\ref{section:Appendix_Universal_Controller_Analysis}.

\textbf{Legged robot control.}
Learning-based locomotion has enabled robust deployment of legged robots through large-scale simulation, system identification, actuator modeling, domain randomization, and policy optimization~\cite{tan2018sim,hwangbo2019learning,kumar2021rma,rudin2022learning}. Additional methods improve robustness and versatility through motion imitation, online adaptation, and explicit adaptation to variations in terrain and system dynamics~\cite{peng2020learning,kumar2021rma}. These advances primarily optimize locomotion for a fixed embodiment, where the observation representation, action parameterization, nominal configuration, and low-level control interface are designed for a specific robot. In contrast, generalized morphology control introduces an additional requirement: a single policy must preserve effective state aggregation and action decoding across multiple bodies with distinct mechanical structures. Our quadruped evaluation studies this cross-platform setting by training shared policies across Go1, Go2, ANYmal-B, and ANYmal-C and comparing them with both generalized morphology baselines and independently trained platform-specific controllers.

\section{Topology-Guided Spatial Recurrence}
\label{sec:spatial_recurrence}

\begin{figure}[t]
\centering
\includegraphics[width=\linewidth]{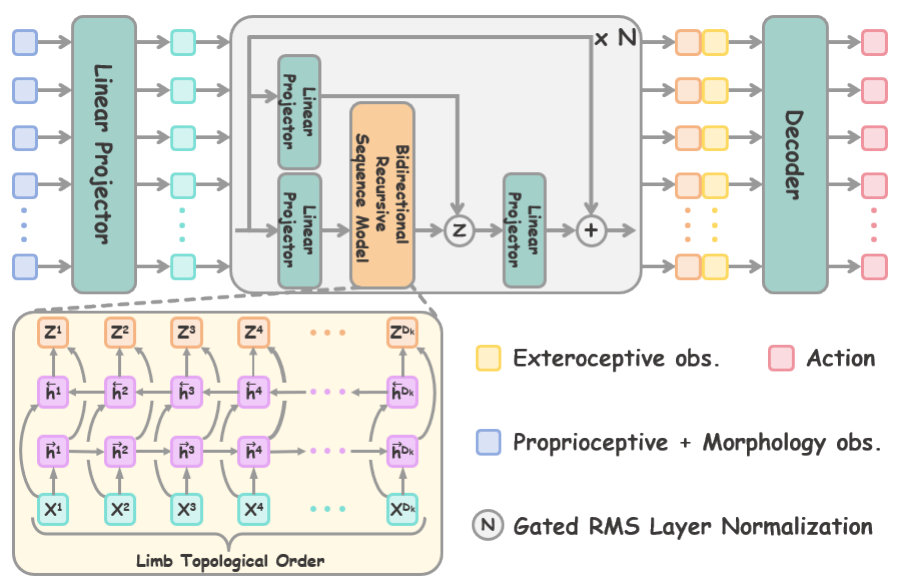}
\caption{RecMorph architecture. A morphology-derived traversal defines the
within-step communication order, and stabilized bidirectional recurrence propagates
whole-body context before shared action decoding.}
\label{fig:architecture}
\vspace{-0.2in}
\end{figure}

\subsection{Problem Formulation}

We consider a family of robot morphologies indexed by $k$, where morphology
$k$ is represented by a rooted kinematic tree
$G_k=(V_k,E_k)$ with $D_k$ limb tokens. At control step $t$, limb $i$ is
described by local proprioception $o_{k,t}^{i}$ and available morphology
attributes $m_k^{i}$. A shared policy $\pi_\theta$ produces actions for the
active actuators, while padded tokens are masked from action probabilities,
entropy, and value aggregation. The learning objective is

\begin{equation}
\max_{\theta}\;
\mathbb{E}_{k\sim p(k),\,\pi_\theta}
\left[
\sum_{t\geq0}\gamma^t r_k(s_t,a_t)
\right],
\label{eq:objective}
\end{equation}

where robot-specific dynamics and rewards are retained while the policy
parameters are shared across morphologies.

To expose the policy to morphology structure, we define a deterministic DFS
preorder $p_k$ over $G_k$ and construct

\begin{equation}
x_j^0=
\sqrt{d}\,
E\!\left[
o_{k,t}^{\,p_k(j)},
m_k^{\,p_k(j)}
\right],
\label{eq:embedding}
\end{equation}

where $E(\cdot)$ is the shared token encoder. The traversal determines the
spatial communication order, while the inverse permutation restores decoded
actions to the physical actuator ordering. Terrain-aware UNIMAL tasks
additionally encode the local height field and fuse it with contextualized
limb features before decoding. Actor and critic use separate parameters, and
the critic aggregates values only over active limbs. The primary formulation assumes a rooted kinematic tree; an extension to
closed-loop morphologies using spanning-tree serialization is evaluated in
Appendix~\ref{sec:closed-loop-kinematic-structures}.

\begin{figure*}[t]
\centering
\includegraphics[width=.92\textwidth]{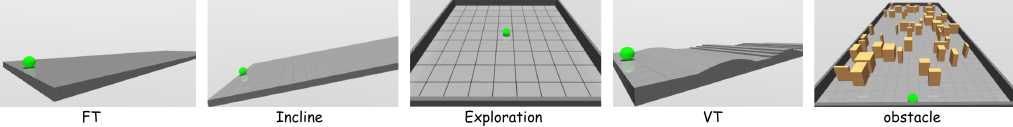}
\caption{UNIMAL evaluation tasks: flat terrain, incline, exploration,
variable terrain, and obstacle traversal.}
\label{fig:environments}
\vspace{-0.2in}
\end{figure*}

\subsection{Stabilized Bidirectional Spatial Recurrence}

Topology-guided serialization determines where information is propagated;
the recurrent block determines how that information is transformed and
integrated at each limb. A useful spatial communication operator must
therefore propagate body-wide context without overwhelming the
limb-specific representation that ultimately drives action decoding.
RecMorph realizes this principle through a stabilized bidirectional
recurrent block that combines normalized feature transformation,
state-dependent modulation, and residual information preservation.

For the primary BiRNN instantiation, each block first applies
RMSNorm~\cite{zhang2019root},
\begin{equation}
N(v)
=
w\odot
\frac{v}{
\sqrt{\operatorname{mean}(v^2)+\epsilon}
},
\label{eq:norm}
\end{equation}
where each normalization layer maintains an independent learnable scale.
Given the representation $x_j^\ell$ of limb token $j$ at block $\ell$,
two parallel projections construct the recurrent input and a
state-dependent modulation signal:
\begin{equation}
u_j
=
\operatorname{SiLU}
\!\left(
W_x N_x(x_j^\ell)
\right),
\qquad
z_j
=
W_z x_j^\ell .
\label{eq:input_projection}
\end{equation}

The recurrent input is propagated in both directions along the
morphology-derived sequence,
\begin{align}
\overrightarrow{h}_j
&=
\tanh\!\left(
A_{\rightarrow}\overrightarrow{h}_{j-1}
+
B_{\rightarrow}u_j
+
b_{\rightarrow}
\right),
\label{eq:forward_rnn}
\\
\overleftarrow{h}_j
&=
\tanh\!\left(
A_{\leftarrow}\overleftarrow{h}_{j+1}
+
B_{\leftarrow}u_j
+
b_{\leftarrow}
\right).
\label{eq:backward_rnn}
\end{align}
The corresponding boundary conditions are
\begin{equation}
\overrightarrow{h}_0=\mathbf{0},
\qquad
\overleftarrow{h}_{D_k+1}=\mathbf{0}.
\label{eq:rnn_boundary}
\end{equation}
The boundary states are reinitialized at every policy invocation, so the
recurrent state represents spatial communication within the current body
configuration rather than temporal memory across control steps. The two
directions are concatenated as
\begin{equation}
y_j
=
\left[
\overrightarrow{h}_j;
\overleftarrow{h}_j
\right],
\label{eq:bidirectional_context}
\end{equation}
providing each limb with context from both sides of the serialized
morphology.

RecMorph then conditions the propagated context on the current limb
representation before residual integration. We first define the
limb-conditioned modulation vector as
\begin{equation}
\tilde g_j
=
\operatorname{SiLU}(z_j),
\qquad
g_j
=
\left[
\tilde g_j;
\tilde g_j
\right].
\label{eq:modulation}
\end{equation}
The duplicated modulation vector matches the dimensionality of the
concatenated forward and backward recurrent representation. The block
output is then
\begin{equation}
x_j^{\ell+1}
=
N_o\!\left(
x_j^\ell
+
W_o
N_g\!\left(
y_j\odot g_j
\right)
\right).
\label{eq:block}
\end{equation}

This construction separates contextual transport from limb-conditioned contextual modulation. Bidirectional recurrence aggregates information across the
body, whereas channel-wise modulation determines which components of that
context are emphasized for the current limb. Applying modulation before
normalization preserves its effect on the relative composition of the
contextual representation, while the residual pathway retains a direct
limb-local signal throughout the recurrent stack. RMS normalization further
controls the scale of repeatedly transformed features, yielding a stable
interface between successive spatial communication blocks. The resulting
block can therefore be viewed as a structured contextualization operator:
topology determines the communication order, recurrence transports
information along that order, and modulated residual integration determines
how the transported context modifies each limb-local representation. This
decomposition is important for generalized morphology control, where the same
communication mechanism must remain applicable across bodies with
different numbers and arrangements of limbs.

The primary UNIMAL configuration stacks four such blocks with an embedding
width of $128$ and a hidden width of $256$ in each direction. Variants such as BiLSTM and BiGRU modify only the recurrent transition while preserving the
normalization, modulation, residual pathway, and decoder interface. This
controlled substitution isolates the choice of recurrent operator from
the stabilization architecture.

\subsection{Topology-Guided Feature Transport}

The serialized morphology turns recurrent propagation into a structured
within-step transport process. To expose the recurrent transport core,
consider a linearized form in which normalization, modulation, residual
integration, and bias terms are temporarily omitted. Partition the
bidirectional output projection as
$W_o=[W_{o,\rightarrow}\;W_{o,\leftarrow}]$. The contextual contribution
at token $i$ can then be written as
\begin{equation}
\Delta x_i =
W_{o,\rightarrow}
\sum_{j=1}^{i}
A_{\rightarrow}^{\,i-j}B_{\rightarrow}u_j
+
W_{o,\leftarrow}
\sum_{j=i}^{D_k}
A_{\leftarrow}^{\,j-i}B_{\leftarrow}u_j .
\label{eq:linearized_transport}
\end{equation}

Hence, information from token $j$ reaches token $i$ through repeated
applications of a shared recurrent transition, with the number of
applications determined by their separation in the serialized morphology.
In the nonlinear controller, the influence is governed by
products of state-dependent transition Jacobians. RecMorph realizes cross-limb contextualization through this progressive
source-to-target transformation. A source feature is repeatedly transformed
as it traverses the morphology-derived sequence, so its contribution to a
target limb depends on the learned transitions and the portion of the body
through which it propagates.

DFS keeps limbs within the same subtree contiguous and places many joints along an articulation chain near one another in the serialized order. Bidirectional propagation therefore supports efficient exchange within kinematic branches while still exposing each token to context from both sides of the body-derived sequence.

The traversal is not unique: different sibling orders preserve the same physical morphology while changing sequence adjacency. For a permutation $P$ and token-to-action network $F_\theta$, the action mean in physical actuator order is

\begin{equation}
\mu_{\theta,P}(X)
=
P^\top F_\theta(PX),
\label{eq:permutation}
\end{equation}

with the same reindexing applied to masks and morphology attributes. During augmentation, children of each parent are independently permuted once per episode and the order is applied consistently throughout the rollout. This exposes the policy to multiple topology-preserving serializations while maintaining a token-to-actuator correspondence. 

We evaluate globally randomized token orders, which disrupt local subtree structure and provide a stronger intervention on the communication path. At fixed hidden width $h$ and block count $L$, recurrent computation scales as $O(LD_kh^2)$, with state storage linear in the number of limb tokens. The experiments in Sec.~\ref{sec:experiments} examine how the
morphology-derived order affects control performance, ordering robustness, throughput, and action relevance. A linearized derivation of the sequence-distance-dependent recurrent
transport operator is provided in Appendix~\ref{app:linearized_transport}.

\begin{figure*}[t]
\centering
\includegraphics[width=\linewidth]{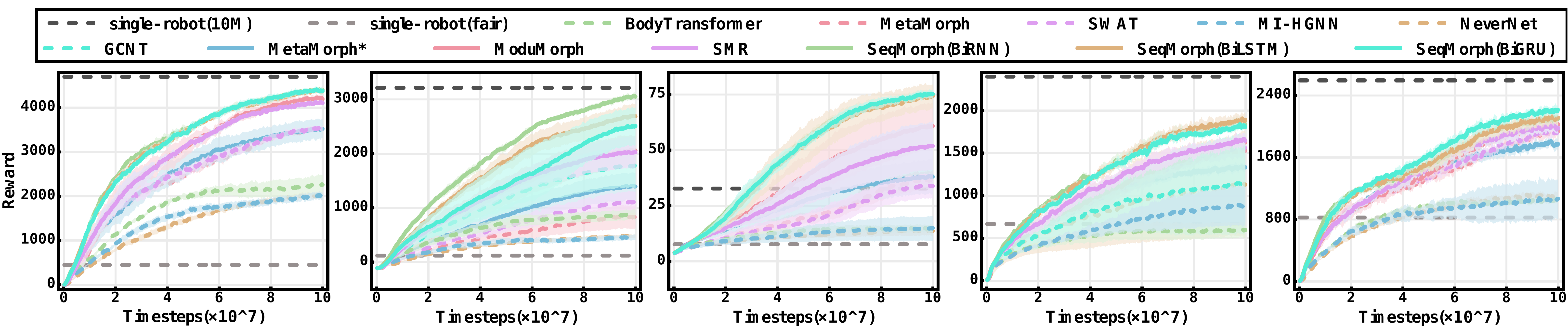}
\caption{Performance of representative generalized morphology controllers across
the five UNIMAL tasks.}
\label{fig:average_rewards_training_ICRA}
\vspace{-0.2in}
\end{figure*}

\section{Generalized Morphology Control}
\label{sec:experiments}

This section evaluates RecMorph on UNIMAL from two complementary
perspectives: its ability to combine strong task performance, high
inference efficiency, zero-shot generalization, and scaling to larger
morphologies; and the roles of recurrent stabilization, morphology-derived
ordering, and cross-limb transformation in producing these gains.

\subsection{Benchmark and Training Protocol}
We evaluate RecMorph on 100 UNIMAL morphologies with up to 12 limbs in MuJoCo~\cite{gupta2021embodied,todorov2012mujoco}. The five tasks in Fig.~\ref{fig:environments} span flat-terrain locomotion (FT), incline traversal, exploration, obstacle traversal, and locomotion over curved slopes, steps, and rugged surfaces (varied terrain, VT). A single generalized morphology policy is trained across all bodies within each task. Terrain-aware tasks additionally provide local height-field observations. 

Policies are trained with PPO~\cite{schulman2017proximal} for approximately 100 million environment interactions. We report online training return as mean $\pm$ sample standard deviation over four independent policy seeds. We compare against representative generalized morphology controllers spanning global attention with MetaMorph ~\cite{gupta2022metamorph} and MetaMorph*~\cite{xiong2023universal}, structure-aware attention with SWAT~\cite{hong2022structure}, contextual modulation with ModuMorph~\cite{xiong2023universal}, graph communication with NerveNet~\cite{wang2018nervenet} and GCNT~\cite{luo2025gcnt}, embodiment-aware attention with Body Transformer (BoT)~\cite{sferrazza2024body}, morphology-informed heterogeneous graph communication with MI-HGNN~\cite{butterfield2025mi}, and shared modular recurrence with SMR~\cite{engwegen2025modular}. 
We also include two single-robot baselines. The single-robot (fair) baseline trains an independent MLP policy for each morphology using the same per-robot training budget as the multi-morphology setting. The single-robot (10M) baseline trains an independent MLP policy for each morphology for 10 million steps and serves as an approximate single-morphology upper reference. 
For BoT and MI-HGNN, we construct policy adaptations compatible with the common generalized morphology control interface.

\subsection{Control Performance and Efficiency}

As shown in Fig.~\ref{fig:average_rewards_training_ICRA}, the RecMorph variant attains the highest mean on every task. Relative to the strongest non-RecMorph mean in each column, the gains are 4.9\% on FT, 48.1\% on incline, 9.1\% on obstacles, 23.7\% on exploration, and 15.1\% on VT. BiRNN leads FT and incline, while BiGRU leads obstacles and exploration, and BiLSTM leads VT. The recurrent transition is therefore a selectable component within the
same spatial-transport architecture. The comparable performance of BiRNN, BiLSTM, and BiGRU further indicates
that the benefit is associated with the shared spatial-transport design
rather than a particular recurrent cell.

\begin{figure}[t]\centering
\includegraphics[width=0.8\linewidth]{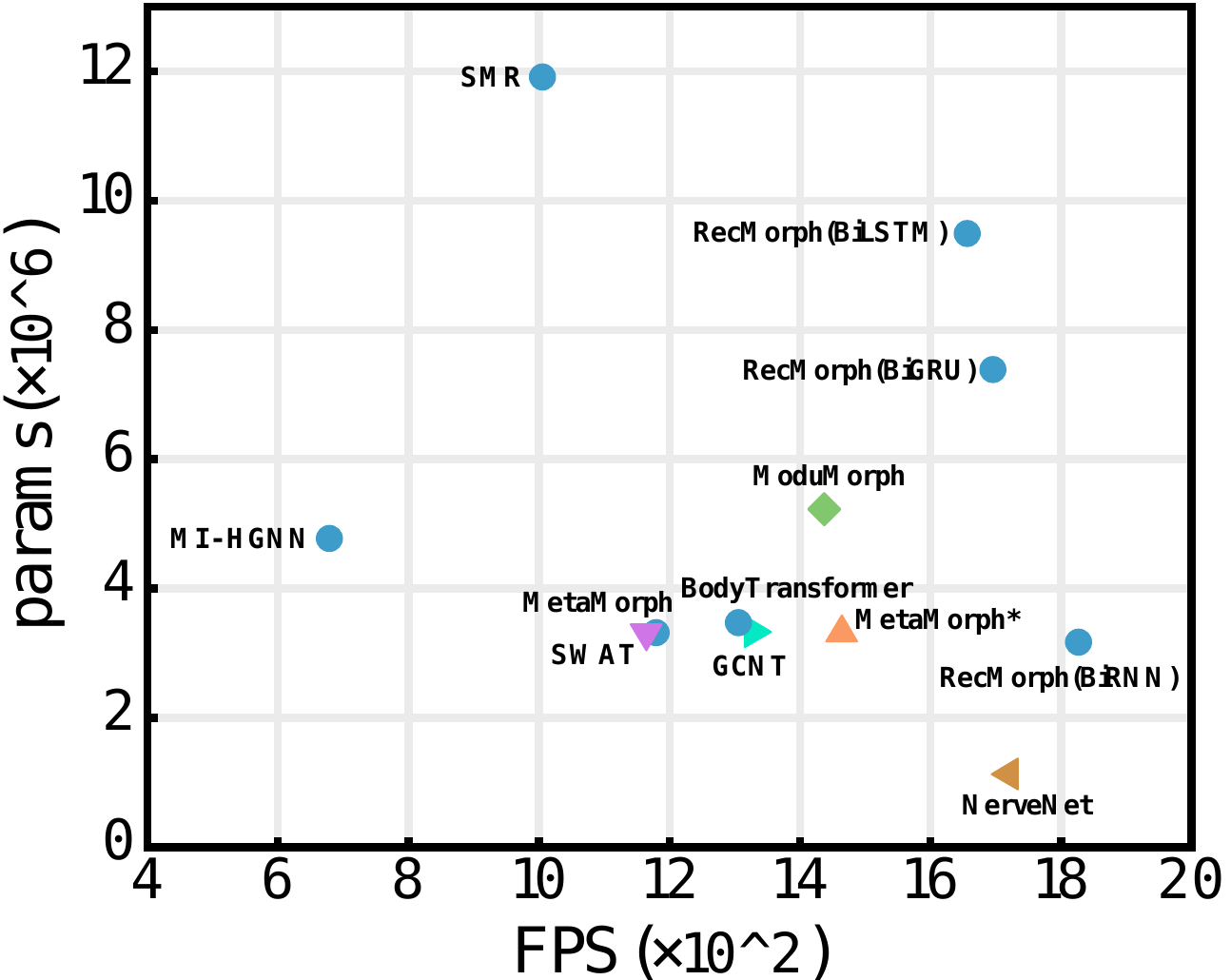}
\caption{Analysis of the number of parameters (params) and frames per second (FPS) for different methods on FT.}
\label{fig:param_time_ICRA}
\vspace{-0.1in}
\end{figure}

In the comparative experiments with baselines, RecMorph employed a four-layer bidirectional sequential model. Figure~\ref{fig:param_time_ICRA} compares model size and measured inference
throughput on FT. RecMorph(BiRNN) achieves the highest FPS among the
evaluated controllers while remaining within a comparable parameter
budget, indicating that recurrent spatial transport provides favorable
runtime scaling in addition to its control-performance gains. Taken together, these results place RecMorph at the intended
performance--efficiency operating point: recurrent whole-body communication
achieves the strongest evaluated control performance together with the
highest measured inference throughput.

\subsection{Zero-Shot Generalization}

\begin{figure}[t]\centering
\includegraphics[width=1\linewidth]{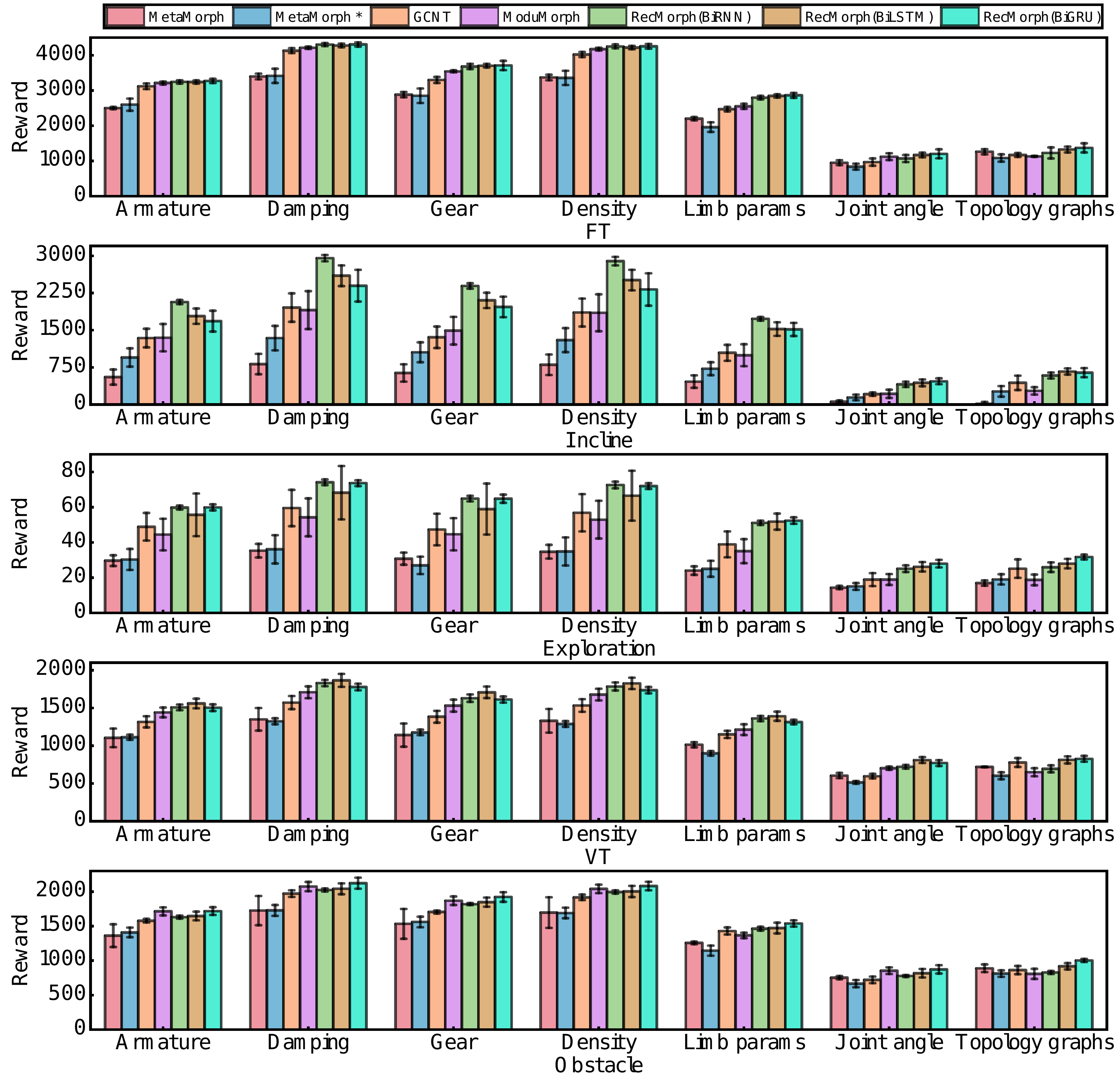}
\caption{Zero-shot transfer under dynamics, kinematic, and topology changes (four policy seeds; bars show mean and error bars show seed SD).}
\label{fig:zeroshot}
\vspace{-0.2in}
\end{figure}

We evaluate whether the learned policy transfers to unseen morphology variations without additional training. As shown in Fig.~\ref{fig:zeroshot}, across these zero-shot perturbations, RecMorph consistently retains strong transfer performance relative to the evaluated generalized morphology baselines. The advantage extends from dynamics and kinematics changes to previously unseen topology graphs, showing that the learned spatial communication rule generalizes beyond the exact training embodiments. These results support topology-guided recurrence as a transferable structural prior rather than a mechanism specialized to a fixed morphology collection. A complementary stress test under single-limb observation dropout is
reported in Appendix~\ref{app:sensor_dropout}.

\subsection{Scaling to Larger Bodies}
\begin{figure}[h]\centering
\includegraphics[width=0.73\linewidth]{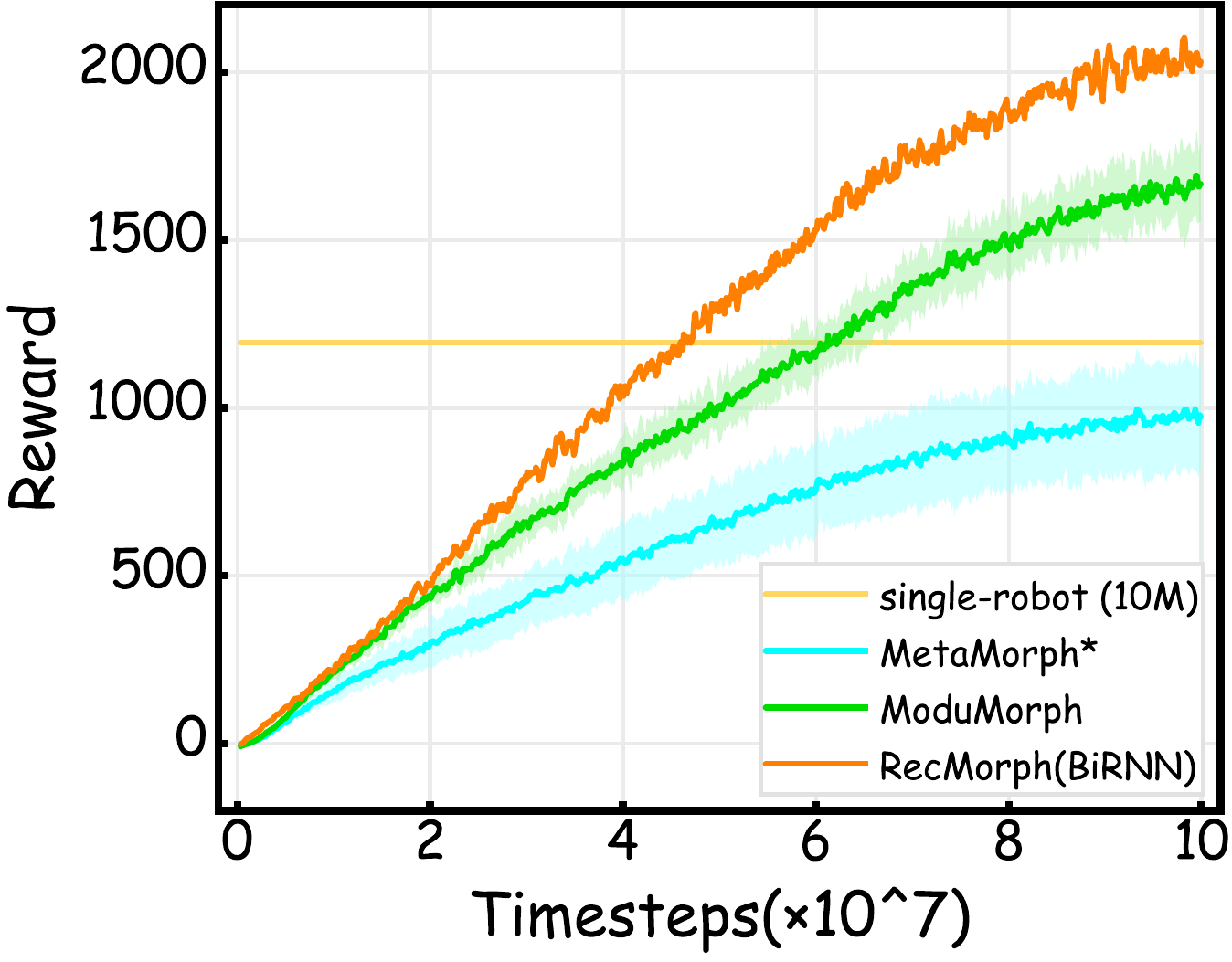}
\caption{FT learning on larger morphologies with up to 30 limbs.
The single-robot reference is independently trained for 10M steps per body.}
\label{fig:larger}
\vspace{-0.1in}
\end{figure}

Larger bodies test whether a shared transition remains useful when both the communication sequence and the action space grow. The larger-body setting increases the maximum limb count from 12 to 30 and the average from approximately 10 to 25. Figure~\ref{fig:larger} shows RecMorph learning effective locomotion in this setting and achieving a higher training-return curve than the attention-based controllers. The single-robot reference is an independently trained MLP with a 10-million-step budget per body; its horizontal line is a final-return reference, not a learning curve. Returns also decrease for the independent MLP reference, showing that larger bodies increase control difficulty beyond the communication operator. Together with zero-shot transfer, this study tests both adaptation-free reuse and shared-policy learning over longer body descriptions. Additional throughput measurements and reference-baseline statistics for
the larger-body setting are reported in
Appendix~\ref{app:larger_morphologies}.

\subsection{Stabilizing Repeated Spatial Transformation}

The recurrent operator repeatedly transforms information before it reaches
later limbs, making preservation of local state and feature scale central
to the architecture. We therefore isolate the components introduced to
stabilize this repeated spatial transformation.

\begin{table}[h]
\centering
\small
\caption{Component ablation. ``Param.-matched'' denotes a plain BiRNN with a parameter count matched to RecMorph.}
\label{tab:ablation}
\setlength{\tabcolsep}{4pt}
\resizebox{\linewidth}{!}{%
\begin{tabular}{lcccc}
\toprule
Variant & Residual & RMSNorm & Channel Modulation & Return \\
\midrule
Plain BiRNN
& -- & -- & -- & $2188.6 \pm 418.7$ \\

Param.-matched
& -- & -- & -- & $2785.2 \pm 488.5$ \\

+ Residual
& \cmark & -- & -- & $3577.0 \pm 260.7$ \\

+ RMSNorm
& \cmark & \cmark & -- & $4027.7 \pm 110.6$ \\

Full RecMorph
& \cmark & \cmark & \cmark & $\mathbf{4367.4 \pm 71.6}$ \\
\bottomrule
\end{tabular}%
}
\vspace{-0.1in}
\end{table}

Table~\ref{tab:ablation} shows a progressive improvement from ordinary recurrence to the full controller. The residual path increases return from 2188.6 to 3577.0, and adding RMSNorm raises it to 4027.7. Channel Modulation provides a further 8.4\% return improvement and 17.0\% learning-curve-area improvement over residual+RMSNorm. This supports a complementary role for information preservation, normalization, and input-dependent channel modulation. The nested interventions quantify each component's effect conditional on the preceding design. The parameter-matched plain model reaches 2785.2 with 3.164 million actor--critic parameters, compared with RecMorph's 3.174 million. A same-V100 diagnostic gives comparable training throughput, 700 versus 698 FPS. The stabilized controller therefore achieves higher return at comparable capacity and measured throughput.

\subsection{Role of Morphology-Derived Order}

Topology provides RecMorph with more than token identity: it determines
the order in which source information is transformed before reaching a
target. We therefore vary the traversal while holding the recurrent
operator fixed.

\begin{figure}[h]\centering
\includegraphics[width=0.73\linewidth]{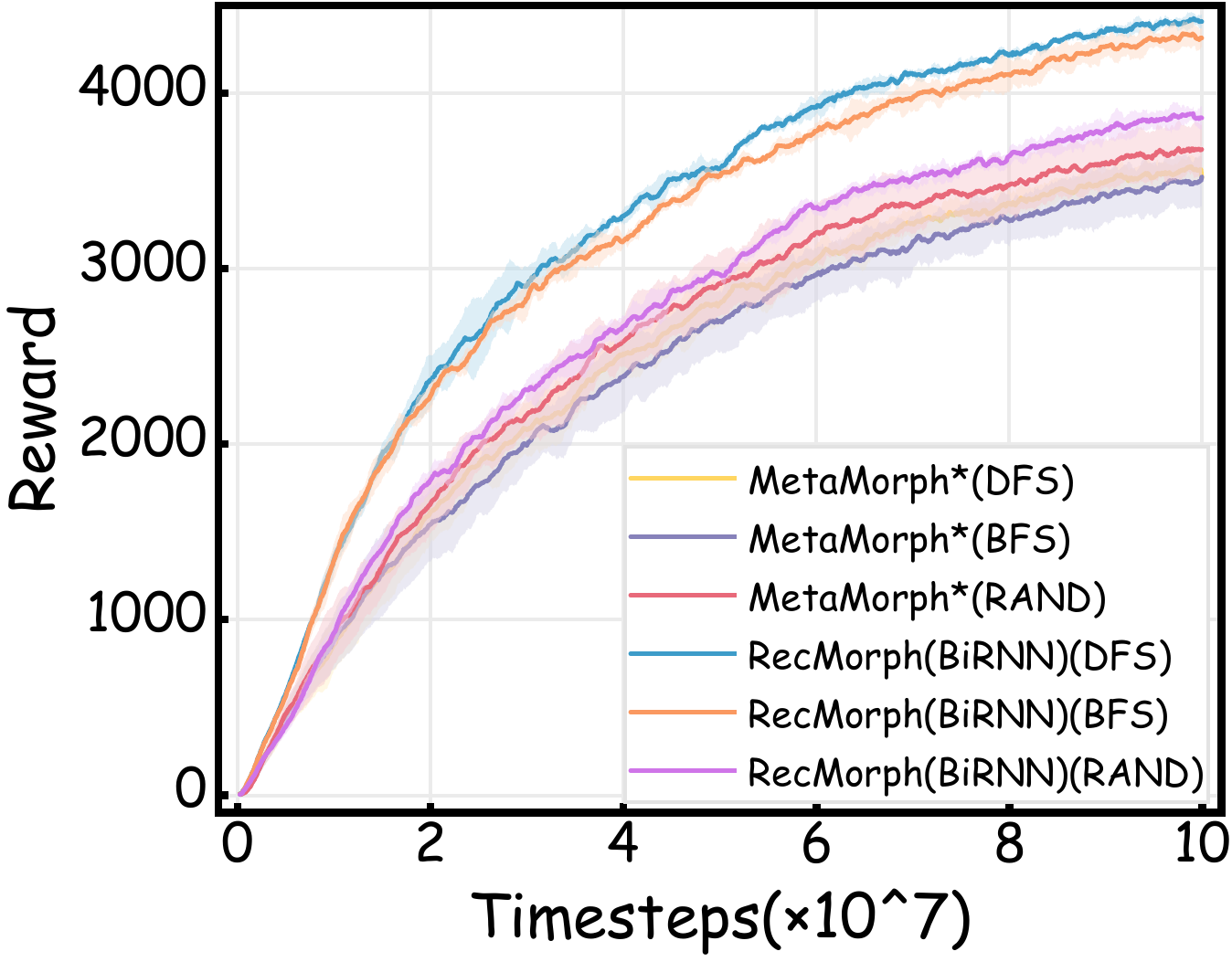}
\caption{Effect of morphology-derived traversal order.
DFS and BFS preserve ordering, whereas global randomization
degrades RecMorph learning.}
\label{fig:traversal}
\vspace{-0.2in}
\end{figure}
Figure~\ref{fig:traversal} compares the effects of depth-first search (DFS), breadth-first search (BFS), and global random ordering on RecMorph. Both tree-derived traversals give stronger RecMorph learning curves than global random order, with DFS leading BFS. The relevant prior is therefore the organization of communication by body structure. Breadth-first order groups limbs by depth, while DFS keeps subtrees contiguous; either retains regularities that an arbitrary token sequence disrupts. DFS supplies a communication order, but a tree admits multiple equivalent sibling orders. We evaluate this choice while keeping the physical robot and actuator mapping fixed. Each of 100 robots is tested under 50 unseen sibling permutations; augmentation samples an equivalent order once per training episode.

\begin{table}[h]\centering\small
\caption{Sibling-order robustness. C: canonical; P: permuted.}
\label{tab:sibling}
\begin{tabular*}{\linewidth}{@{\extracolsep{\fill}}lrrrr@{}}\toprule
Train / seed & C test & P test & Drop (\%) & Tail\\\midrule
C / 1 &1095.0&399.3&63.5&319.1\\
C / 1409&1064.8&413.8&61.1&319.3\\
P / 1&1369.3&1268.1&7.4&1164.4\\
P / 1409&1138.3&1094.0&3.9&993.7\\\bottomrule
\end{tabular*}
\end{table}

Table~\ref{tab:sibling} reveals both the strength and flexibility of the sequence prior. Canonical policies depend strongly on their training order, losing 61.1--63.5\% under sibling permutations. Episode-level augmentation reduces the loss to 3.9--7.4\% and improves the worst-decile return. Canonical evaluation also improves in these two runs. The intervention therefore offers a practical way to make the learned controller less dependent on an arbitrary serialization choice. These results measure learned ordering robustness using episode evaluation, separately from the online training benchmark.

\subsection{Cross-Limb Transformation and Action Relevance}
\label{subsec:representation_alignment}

We examine how recurrent spatial transport changes the compatibility of
limb representations with a shared action decoder. Matched linear probes
are fitted before and after communication on four independently trained
policies.

\begin{table}[h]
\centering
\small
\caption{Linear probes before and after recurrent communication.}
\label{tab:probes}
\setlength{\tabcolsep}{2pt}
\begin{tabular}{lccc}
\toprule
Probe & Pre & Post & $\Delta$ \\
\midrule
Robot-ID acc. 
& $84.28 \pm 1.26$
& $61.46 \pm 3.19$
& $\mathbf{-22.82 \pm 3.63}$ \\

Limb-ID acc. 
& $95.49 \pm 0.39$
& $86.45 \pm 2.62$
& $\mathbf{-9.05 \pm 2.44}$ \\

Action $R^2$
& $-0.071 \pm 0.026$
& $0.354 \pm 0.021$
& $\mathbf{+0.425 \pm 0.028}$ \\

Action NRMSE
& $1.033 \pm 0.013$
& $0.802 \pm 0.014$
& $\mathbf{-0.231 \pm 0.015}$ \\
\bottomrule
\end{tabular}
\end{table}

As shown in Table~\ref{tab:probes}, communication substantially reduces
explicit morphology identity while improving action predictability. Robot-ID
and limb-ID balanced accuracy decrease from $84.28\pm1.26$ to
$61.46\pm3.19$ and from $95.49\pm0.39$ to $86.45\pm2.62$, respectively.
Meanwhile, action $R^2$ increases from $-0.071\pm0.026$ to
$0.354\pm0.021$, and NRMSE decreases from $1.033\pm0.013$ to
$0.802\pm0.014$. Together, the probes show that recurrent transport reduces explicit
source-identity information while producing representations that are more
predictive of the policy action. This behavior is consistent with the
intended cross-limb transformation. Complementary feature-space
analysis is provided in Appendix~\ref{app:feature_space}.

\section{Cross-Platform Generalized Control}

The UNIMAL experiments establish the performance, efficiency, and
generalization of RecMorph under a unified procedural morphology and
control interface. We extend this evaluation to standard quadruped
platforms, where controller architectures must operate across different
body dynamics and low-level control conventions. A matched benchmark
spanning Go1, Go2, ANYmal-B, and ANYmal-C evaluates whether the advantage
of RecMorph persists beyond the UNIMAL setting, followed by physical
deployment of one shared policy on Go1 and Go2.

\begin{figure*}[t]\centering
\vspace{-0.2in}
\includegraphics[width=0.7\textwidth]{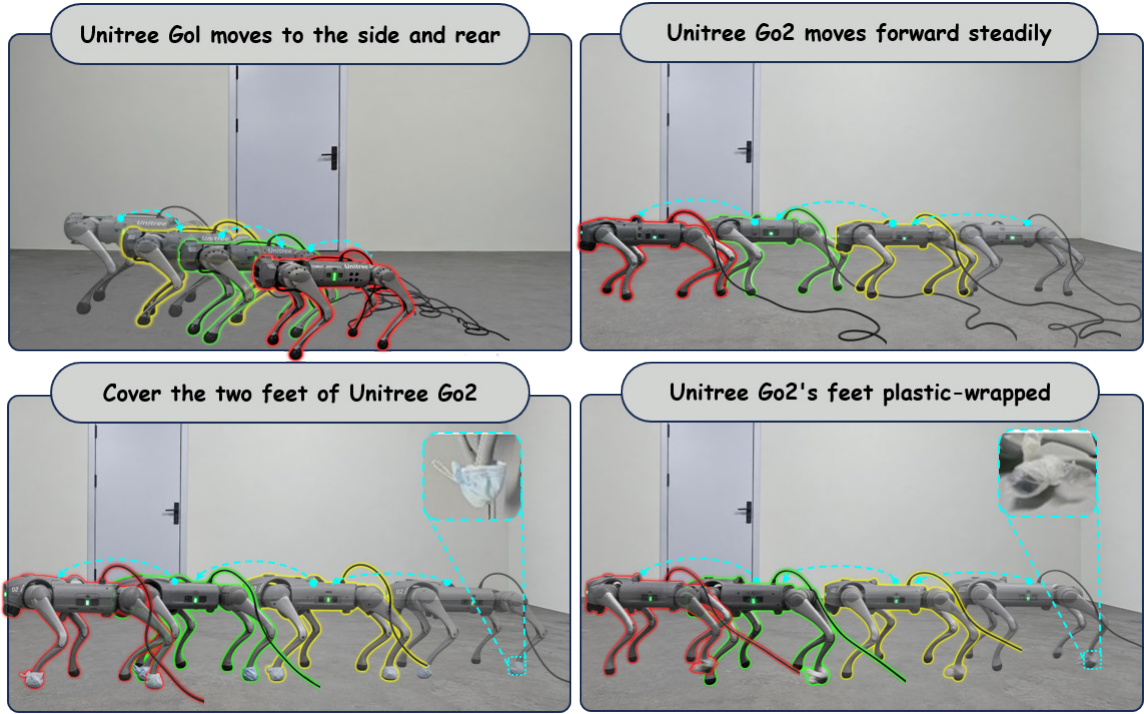}
\caption{Shared-policy deployment on physical Go1 and Go2 robots. All four conditions use the same RecMorph policy}
\label{fig:hardware}
\end{figure*}
\subsection{Transfer to Four Quadruped Platforms}

We transfer RecMorph, MetaMorph, ModuMorph, BodyTransformer, and GCNT to
the same four-platform benchmark, with each method sharing one policy
across Go1, Go2, ANYmal-B, and ANYmal-C. Each robot is represented through
a common limb-token observation and action interface. A platform adapter
maps native joint states into the shared token order and maps the active
policy outputs back to the corresponding native joints. Robot-specific
reference poses, low-level gains, joint limits, and safety constraints
remain in the actuation layer. The resulting interface allows the same
generalized controller architecture to operate across all four platforms
while preserving the low-level control settings required by each robot. The specialist MLP baseline is trained independently for each robot,
whereas all generalized-controller baselines share one policy across the
four platforms. All methods use 128 environments per robot, 32 rollout steps, and
10,000 PPO iterations. Four independent random seeds are used for every
method.

\begin{table*}[t]
    \centering
    \caption{Nominal-friction quadruped benchmark.}
    \label{tab:friction}
    \setlength{\tabcolsep}{9pt}
    \renewcommand{\arraystretch}{1.08}
    \resizebox{0.85\textwidth}{!}{%
    \begin{tabular}{lcrrrr}
        \toprule
        Method & Controller scope
        & Velocity RMSE (m/s) $\downarrow$
        & Progress (m) $\uparrow$
        & Tilt RMS ($^{\circ}$) $\downarrow$
        & Fall rate (\%) $\downarrow$ \\
        \midrule
        \textbf{RecMorph(BiRNN)}
        & Shared
        & $\mathbf{0.156 \pm 0.038}$
        & $\mathbf{97.11 \pm 0.82}$
        & $\mathbf{3.00 \pm 0.62}$
        & $\mathbf{6.97 \pm 11.11}$ \\

        Per-robot MLP
        & Specialist
        & $0.276 \pm 0.116$
        & $92.15 \pm 7.06$
        & $5.52 \pm 0.55$
        & $17.04 \pm 11.47$ \\

        ModuMorph
        & Shared
        & $0.648 \pm 0.585$
        & $48.49 \pm 58.70$
        & $3.61 \pm 2.17$
        & $43.35 \pm 9.41$ \\

        MetaMorph
        & Shared
        & $0.779 \pm 0.016$
        & $29.27 \pm 0.65$
        & $5.14 \pm 0.32$
        & $34.26 \pm 0.78$ \\

        BodyTransformer
        & Shared
        & $0.965 \pm 0.143$
        & $15.81 \pm 8.85$
        & $16.54 \pm 0.56$
        & $96.69 \pm 3.25$ \\

        GCNT
        & Shared
        & $1.174 \pm 0.160$
        & $-6.00 \pm 10.72$
        & $17.11 \pm 4.04$
        & $99.94 \pm 0.08$ \\
        \bottomrule
    \end{tabular}%
    }
    \vspace{-0.1in}

\end{table*}

Table~\ref{tab:friction} reports four-platform macro results under nominal static/dynamic friction $(0.8,0.6)$. RecMorph provides the best tracking, progress, stability, and fall-rate trade-off among shared and specialist controllers. Relative to independently trained per-robot MLPs, RecMorph reduces velocity RMSE from $0.276\pm0.116$ to $0.156\pm0.038$\,m/s (43.5\%), reduces tilt RMS from $(5.52\pm0.55)^{\circ}$ to $(3.00\pm0.62)^{\circ}$ (45.6\%), and lowers the fall rate from $17.04\pm11.47$\% to $6.97\pm11.11$\% (59.1\%), while increasing progress from $92.15\pm7.06$ to $97.11\pm0.82$\,m (5.4\%). RecMorph has the best mean on all four metrics among the shared-policy baselines. Relative to ModuMorph, the strongest shared controller in nominal velocity tracking, RecMorph reduces mean RMSE by 75.9\% and approximately doubles mean progress.

Table~\ref{tab:friction_all} compares all six controllers under low, nominal, and high friction. Every entry reports the mean and sample SD across independent policy seeds. Static/dynamic coefficients are $(0.3,0.2)$, $(0.8,0.6)$, and $(1.2,1.0)$, respectively.

Across the four evaluated platforms, RecMorph achieves the best macro-averaged performance on all four metrics under both nominal and high friction, outperforming the shared-controller baselines and independently trained specialist MLPs. Under high friction, it achieves $0.133\pm0.023$\,m/s velocity RMSE, $98.40\pm1.97$\,m progress, $(2.69\pm0.85)^{\circ}$ tilt RMS, and a $(6.74\pm12.26)\%$ fall rate using a single shared policy. These results demonstrate that the aggregate advantage of RecMorph extends across two contact settings, supporting effective policy sharing across the evaluated embodiments without requiring separate platform-specific policies. Under low friction, RecMorph retains advantages over several shared-controller baselines, achieving lower mean velocity RMSE and greater mean progress than MetaMorph, BodyTransformer, and GCNT. Compared with specialist MLPs, it achieves slightly lower mean velocity RMSE ($0.907\pm0.042$ versus $0.919\pm0.198$\,m/s) and lower mean tilt RMS ($(20.95\pm5.45)^{\circ}$ versus $(29.21\pm9.75)^{\circ}$), with only a modest reduction in mean progress ($12.99\pm3.27$ versus $15.22\pm18.02$\,m). This competitiveness does not extend to fall resistance: RecMorph has a higher fall rate, $(77.14\pm14.24)\%$ versus $(28.25\pm28.17)\%$. ModuMorph achieves better low-friction means on all four metrics, although its observed seed-to-seed SD is substantially larger for velocity RMSE and progress. These results show that the performance advantage of RecMorph persists
when generalized controllers are transferred from procedural UNIMAL
morphologies to standard quadruped platforms, supporting the
transferability of its recurrent cross-limb transformation beyond the
original benchmark setting.

\subsection{Physical Deployment on Go1 and Go2}
We deploy one separately trained shared RecMorph policy on physical Go1 and Go2 to assess whether the high-level mapping can be reused without platform-specific policy retraining. Each condition contains ten 15-s trials, all completed without a reported fall. Detailed deployment training settings and trial-level success statistics
are provided in Appendices~\ref{appendix:quadruped_details}
and~\ref{appendix:hardware_results}.

\begin{table*}[t]
    \centering
    \caption{Cross-platform results under friction variation.}
    \label{tab:friction_all}
    \setlength{\tabcolsep}{8pt}
    \renewcommand{\arraystretch}{1.08}
    \resizebox{0.82\textwidth}{!}{%
    \begin{tabular}{llrrrr}
        \toprule
        Method
        & Friction
        & Velocity RMSE (m/s) $\downarrow$
        & Progress (m) $\uparrow$
        & Tilt RMS ($^{\circ}$) $\downarrow$
        & Fall rate (\%) $\downarrow$ \\
        \midrule

        RecMorph(BiRNN)
        & Low
        & $0.907 \pm 0.042$
        & $12.99 \pm 3.27$
        & $20.95 \pm 5.45$
        & $77.14 \pm 14.24$ \\

        & Nominal
        & $\mathbf{0.156 \pm 0.038}$
        & $\mathbf{97.11 \pm 0.82}$
        & $\mathbf{3.00 \pm 0.62}$
        & $\mathbf{6.97 \pm 11.11}$ \\

        & High
        & $\mathbf{0.133 \pm 0.023}$
        & $\mathbf{98.40 \pm 1.97}$
        & $\mathbf{2.69 \pm 0.85}$
        & $\mathbf{6.74 \pm 12.26}$ \\

        \addlinespace[2pt]
        Per-robot MLP
        & Low
        & $0.919 \pm 0.198$
        & $15.22 \pm 18.02$
        & $29.21 \pm 9.75$
        & $\mathbf{28.25 \pm 28.17}$ \\

        & Nominal
        & $0.276 \pm 0.116$
        & $92.15 \pm 7.06$
        & $5.52 \pm 0.55$
        & $17.04 \pm 11.47$ \\

        & High
        & $0.332 \pm 0.093$
        & $85.32 \pm 3.37$
        & $8.54 \pm 1.24$
        & $25.74 \pm 12.23$ \\

        \addlinespace[2pt]
        ModuMorph
        & Low
        & $\mathbf{0.793 \pm 0.364}$
        & $\mathbf{29.63 \pm 32.89}$
        & $\mathbf{4.52 \pm 2.58}$
        & $49.82 \pm 0.26$ \\

        & Nominal
        & $0.648 \pm 0.585$
        & $48.49 \pm 58.70$
        & $3.61 \pm 2.17$
        & $43.35 \pm 9.41$ \\

        & High
        & $0.654 \pm 0.576$
        & $47.83 \pm 57.75$
        & $3.91 \pm 2.56$
        & $46.49 \pm 4.96$ \\

        \addlinespace[2pt]
        MetaMorph
        & Low
        & $0.995 \pm 0.003$
        & $1.43 \pm 0.26$
        & $5.73 \pm 0.33$
        & $36.09 \pm 17.64$ \\

        & Nominal
        & $0.779 \pm 0.016$
        & $29.27 \pm 0.65$
        & $5.14 \pm 0.32$
        & $34.26 \pm 0.78$ \\

        & High
        & $0.748 \pm 0.008$
        & $31.84 \pm 0.92$
        & $6.62 \pm 0.91$
        & $44.95 \pm 1.19$ \\

        \addlinespace[2pt]
        BodyTransformer
        & Low
        & $1.000 \pm 0.013$
        & $3.83 \pm 1.47$
        & $13.24 \pm 2.03$
        & $94.02 \pm 4.88$ \\

        & Nominal
        & $0.965 \pm 0.143$
        & $15.81 \pm 8.85$
        & $16.54 \pm 0.56$
        & $96.69 \pm 3.25$ \\

        & High
        & $0.972 \pm 0.158$
        & $15.28 \pm 10.04$
        & $16.85 \pm 0.95$
        & $96.89 \pm 2.95$ \\

        \addlinespace[2pt]
        GCNT
        & Low
        & $1.189 \pm 0.211$
        & $-11.58 \pm 16.24$
        & $13.10 \pm 7.47$
        & $99.06 \pm 1.33$ \\

        & Nominal
        & $1.174 \pm 0.160$
        & $-6.00 \pm 10.72$
        & $17.11 \pm 4.04$
        & $99.94 \pm 0.08$ \\

        & High
        & $1.211 \pm 0.093$
        & $-7.08 \pm 4.56$
        & $16.58 \pm 4.73$
        & $99.42 \pm 0.82$ \\
        \bottomrule
    \end{tabular}%
    }
    \vspace{-0.1in}
\end{table*}

The deployed checkpoint uses policy seed 1409 and a 64-token, 32-feature interface with 12 active actuator tokens per quadruped. The hardware experiment demonstrates reuse of a shared policy on two physical platforms. Active tokens contain base velocity, projected gravity, velocity commands, joint states, previous actions, and joint-limit features. The shared decoder produces joint-position offsets according to
\begin{equation}
    q_{\mathrm{target},k}=q_{\mathrm{default},k}+0.25\,a_{\theta,k}.
\end{equation}
The deployment policy runs at 50\,Hz and is trained on a mixture of flat and micro-rough terrain with corrupted proprioceptive observations. Thus, the same learned high-level controller is reused across Go1 and Go2 while the low-level interface respects their physical differences. The deployment therefore evaluates the same shared-policy formulation on
physical hardware without changing the learned policy across platforms. Trial-level deployment statistics and the success criterion are reported
in Appendix~\ref{appendix:hardware_results}.

\section{Discussion}

RecMorph shows that recurrent sequence computation provides a useful
mechanism for Generalized Morphology Control because communication and
cross-limb representation transformation are performed by the same
operator. Robot topology determines the order of transformation, while a
shared recurrent transition progressively converts source-limb information
into representations used by other limb controllers. The comparable
behavior of BiRNN, BiLSTM, and BiGRU indicates that the central mechanism
extends beyond a particular recurrent cell. The quadruped experiments add
a second result: generalized controller architectures developed on
procedural morphology benchmarks can be transferred to standard robot
platforms through a shared token/action interface while retaining
platform-specific low-level actuation.

The experiments also expose concrete limits of the current formulation.
First, RecMorph commits to a fixed topology-consistent traversal, while the
sibling-order study shows that different valid serializations can lead to
different learned transformations. Learning task-dependent traversal
orders or combining several topology-consistent routes is therefore a
natural extension. Second, linear computational complexity does not prevent
the source-to-target transformation path from growing with body size;
hierarchical recurrence over limbs, subtrees, and branches could shorten
this path for substantially larger robots. Third, the low-friction results
separate morphology generalization from dynamics adaptation, motivating
recurrent transformations conditioned on online estimates of contact,
friction, and actuator response. 

\section{Conclusion}

We introduced RecMorph for Generalized Morphology Control, using robot
topology to organize recurrent cross-limb transformation. Shared
bidirectional transitions progressively convert limb-local information
into target-relevant representations, providing whole-body communication
with linear token complexity at fixed network width and depth. UNIMAL
experiments establish the performance, efficiency, and generalization of
this mechanism, while the cross-platform study shows that its advantage
survives the transfer of generalized controllers to standard quadruped
platforms and physical Go1/Go2.

\section*{ACKNOWLEDGMENT}

This work was supported by the National Natural Science Foundation of China
(Grant No.~52472448); the Deep Earth Probe and Mineral Resources Exploration
National Science and Technology Major Project
(Grant Nos.~2024ZD1000804 and 2024ZD1000802); the Scientific and
Technological Research Project of the Education Department of Jilin Province
(Grant No.~JJKH20261598KJ); and the Natural Science Foundation of Jilin
Province (Grant No.~20260205078GH).

\bibliographystyle{IEEEtran}
\bibliography{references}

\newpage
\onecolumn

\begingroup

\setcounter{section}{0}
\setcounter{subsection}{0}

\renewcommand{\thesection}{\Alph{section}}
\renewcommand{\thesubsection}{\thesection.\arabic{subsection}}

\newcommand{\appsection}[1]{%
    \refstepcounter{section}%
    \setcounter{subsection}{0}%
    \vspace{0.8em}%
    \noindent
    {\normalsize\bfseries
    \thesection\hspace{0.8em}#1\par}%
    \vspace{0.5em}%
}

\newcommand{\appsubsection}[1]{%
    \refstepcounter{subsection}%
    \vspace{0.8em}%
    \noindent
    {\normalsize\bfseries
    \thesubsection\hspace{0.7em}#1\par}%
    \vspace{0.35em}%
}

\appsection{Morphology as a Structural Prior for Cross-Limb Representation Compatibility}
\label{sec:Morphology_as_a_Structural_Prior_for_State_Aggregation}

\begin{figure*}[h]  
  \vspace{-0.1in}    
  \centering
  \includegraphics[width=\linewidth]{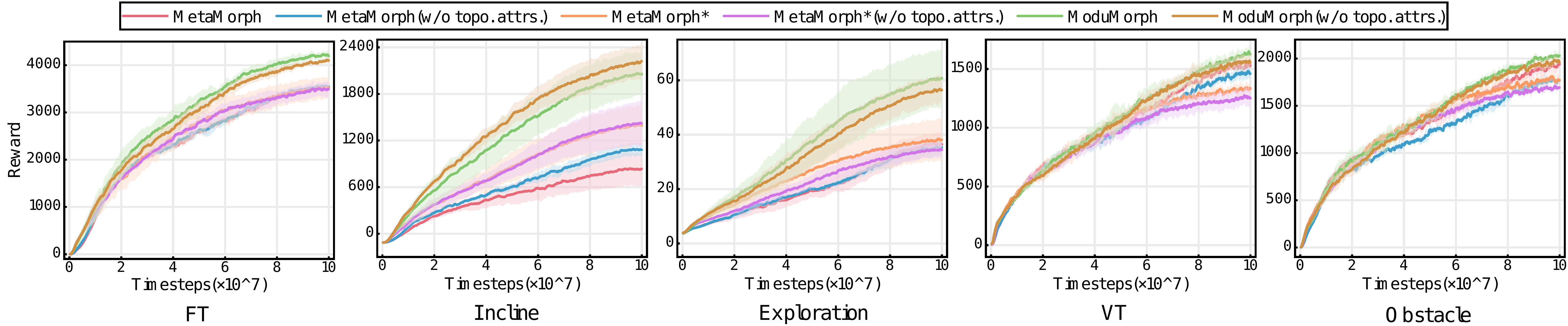} 
  \caption{Performance after removing selected topology-dependent morphology
attributes from the UNIMAL observation space. Topology-derived token
ordering is retained where applicable.}  
  \label{fig:section_3_1}  
  \vspace{-0.2in}    
\end{figure*}

Generalized morphology control requires information expressed in different
limb-local physical contexts to become useful for target-specific action
prediction. We refer to this general computation as
\emph{cross-limb contextualization}: a target representation is constructed
from source-limb information through target-dependent weighting,
morphology-conditioned transformation, progressive propagation, or a
combination of these operations. The resulting ability of heterogeneous
limb representations to support common downstream computation is referred
to as \emph{cross-limb representation compatibility}. MetaMorph, ModuMorph, NerveNet, and RecMorph realize this computation in
different ways. MetaMorph relies primarily on state-dependent global
aggregation, ModuMorph additionally adapts limb projections according to
morphology, NerveNet performs graph-local message transformation, and
RecMorph progressively transforms information along a morphology-derived
sequence.

\appsubsection{Sensitivity to Topology-Dependent Morphology Attributes}

\label{app:topology_attribute_sensitivity}

UNIMAL morphology observations combine hardware descriptors with fields
encoding relative position and orientation. We isolate the contribution of
these explicit topology-dependent attributes by removing the fields listed
in Table~\ref{tab:topology_removal_features} while preserving the remaining
observation interface. Architecture-level structural cues, including
morphology-derived token ordering, are retained. This intervention therefore
measures controller sensitivity to explicit topology-dependent morphology
attributes.

MetaMorph combines DFS-ordered limb tokens with Transformer-based global
communication and a learned positional embedding. We additionally evaluate
MetaMorph*, which removes the positional encoding while retaining the same
token ordering and attention architecture. For both variants, we then remove
the selected topology-dependent morphology attributes defined in
Table~\ref{tab:topology_removal_features}. Figure~\ref{fig:section_3_1}
shows that this attribute intervention does not produce a systematic
decrease in return across the five tasks, indicating limited sensitivity
to these explicit morphology fields under the evaluated protocol.

\begin{equation}
\label{equ:pluspos}
\mathbf{\hat{E} }_{k,t} = \mathbf{E }_{k,t}+\mathbf{W_{pos}}
\end{equation}

ModuMorph introduces morphology-conditioned limb projections and morphology-conditioned attention through a hypernetwork. Under the same
attribute-removal intervention, its return also remains broadly stable
across the evaluated tasks. Fig.~\ref{fig:section_3_1} indicates that the selected
explicit topology-dependent attributes are not the sole source of useful
morphology conditioning in this setting; the remaining morphology
descriptors continue to support body-dependent feature transformation.

\appsubsection{Cross-Limb Representation Compatibility}
\label{app:representation_compatibility}

Limb observations correspond to different mechanical roles across
morphologies, so a shared controller must transform them into
representations that can be jointly compared, propagated, and decoded by
common downstream computation. We refer to this property as
\emph{cross-limb representation compatibility}. Different controller
families realize this compatibility through different computational
mechanisms: shared projection and global attention in MetaMorph,
morphology-conditioned mappings in ModuMorph, graph-local communication
in NerveNet, and morphology-guided recurrent transport in RecMorph.

We first analyze MetaMorph from this perspective. MetaMorph projects each raw limb observation ${\mathbf{S}_{\mathbf{MP}}}_{k,t}^{i}$ into a latent embedding using a shared linear layer. Although this projection maps all limb observations into the same feature dimension, it does not explicitly account for limb-specific coordinate frames or physical roles before aggregation. The subsequent global aggregation is performed by Multi-Head Self-Attention. The output $\mathbf{Z}_{k,t}^{i,(l)}$ for limb $i$ at the $l^{th}$ layer at time $t$ is as seen in Equation \ref{equ:MetaMorph_equ}. Detailed analysis of MetaMorph is provided in Appendix \ref{subsection:Appendix_MetaMorph}.

\begin{equation}
    \mathbf{\hat{\Gamma}}_{k,t}^{i} = \sum_{j \neq i} {\mathbf{R}_{j,i}^{(l)}}^\top\mathbf{Z}_{k,t}^{j,(l-1)}  + \left ( {\mathbf{R}_{i,i}^{(l)}}  +\mathbf{I} \right )^\top\mathbf{Z}_{k,t}^{i,(l-1)}
\end{equation}

\begin{equation}
\label{equ:MetaMorph_equ}  
    \mathbf{Z}_{k,t}^{i,(l)} = \mathbf{W}_{\textbf{MX}}^{\left ( l \right )}{}^\top \mathbf{\hat{\Gamma}}_{k,t}^{i} + \mathbf{b}_\mathbf{O}^{(l)}
\end{equation}

Here, $\mathbf{R}_{j,i}^{(l)}$ denotes the attention-based aggregation weight from limb $j$ to limb $i$. For a given attention head, the value projection of source limb $j$ is
shared across target limbs. Target dependence enters through the attention
coefficients, which determine how strongly the same source
representation contributes to different targets. MetaMorph therefore
performs target-dependent aggregation, while the source-to-target feature
transformation itself remains implicit in the weighted mixing operation.

ModuMorph can be interpreted differently. It uses a hypernetwork to generate limb-wise projection weights $\mathbf{\tilde{W}}_{\mathbf{emb}_{\mathbf{HN}}}^{i}$ and biases $\mathbf{b}_{\mathbf{emb}_{\mathbf{HN}}}^{i}$ from the morphological context $\mathbf{C}_{k}$. The input embedding of limb $i$ is given by

\begin{equation}
\label{equ:Modumorph_encoder_equ}  
 {\mathbf{E}_{\textbf{HN}}}_{k,t}^{i} = \left( {\mathbf{\tilde{W}}_{\textbf{emb}_\textbf{HN}}^{i}}{}^\top {\mathbf{S}_\textbf{MP}}_{k,t}^{i} + \mathbf{b}_{\textbf{emb}_\textbf{HN}}^{i} \right) \cdot \sqrt{D_{\text{emb}}} 
\end{equation}

Compared with a single shared projection, these morphology-conditioned limb-wise projections can make local limb representations more stable and more compatible before aggregation. In this sense, ModuMorph improves cross-limb representation compatibility through morphology-conditioned projections.

ModuMorph also differs from standard self-attention in its aggregation mechanism. Its queries and keys are generated from the morphological context $\mathbf{C}_{k}$, resulting in time-invariant aggregation weights $\mathbf{R}_{\mathbf{HN},j,i}^{(l)}$. For limb $i$, the aggregation can be written as

\begin{equation}
    \mathbf{\tilde{\Gamma}}_{k,t}^{i} = \sum_{j \neq i} {\mathbf{R}_{\textbf{HN},j,i}^{(l)}}{}^\top \mathbf{Z}_{k,t}^{j,(l-1)}  + \left ( {\mathbf{R_{\textbf{HN}}}_{i,i}^{(l)}} +\mathbf{I} \right )^\top \mathbf{Z}_{k,t}^{i,(l-1)}
\end{equation}

\begin{equation}
\label{equ:ModuMorph_equ}
    {\mathbf{Z}_{\textbf{HN} }}_{k,t}^{i,(l)} = \mathbf{W}_{\textbf{MX}}^{\left ( l \right )}{}^\top \mathbf{\tilde{\Gamma}}_{k,t}^{i} + \mathbf{b}_\mathbf{O}^{(l)}
\end{equation}

Because $\mathbf{R}_{\mathbf{HN},j,i}^{(l)}$ is conditioned on morphology rather than recomputed solely from instantaneous observations, Because $\mathbf{R}_{\mathrm{HN},j,i}^{(l)}$ is conditioned on morphology,
ModuMorph produces a morphology-dependent aggregation pattern that remains
fixed for a given body. ModuMorph addresses part of this representation mismatch by generating
morphology-conditioned limb projections. Each source limb can therefore
enter aggregation through a body-dependent feature map. The subsequent
whole-body communication still uses dense attention, so morphology
primarily adapts the local feature transformation while global
source-to-target aggregation remains pairwise.

This analysis leads to our central design perspective. Effective generalized morphology control requires both cross-limb contextualization and body-wide information aggregation. Existing attention-based controllers provide global aggregation, while morphology-conditioned hypernetworks improve the compatibility of limb-local features. RecMorph builds on this perspective by using morphology as a structural prior for recursive topological transport, enabling limb-local information to be progressively transformed and aggregated along the robot's kinematic structure.

\appsection{Universal Controller Analysis}
\label{section:Appendix_Universal_Controller_Analysis}

\appsubsection{MetaMorph}
\label{subsection:Appendix_MetaMorph}

This section analyzes MetaMorph from the perspective of cross-limb contextualization and aggregation. The goal is to characterize how its self-attention operator mixes limb-local features. While MetaMorph includes a positional encoding layer intended for morphological identification, our analysis aligns with recent findings~\cite{xiong2023universal} suggesting that this encoding acts primarily as a context-dependent bias rather than a robust topological descriptor. Hence we omit the positional encoding layer. This configuration is referred to as MetaMorph*. The policy network first projects the raw observation ${\mathbf{S}_\textbf{MP}}_{k,t}^i \in \mathbb{R}^{D_{\text{obs}}}$ of each limb $i$ of robot $k$ at time step $t$ into a latent space $\mathbb{R}^{D_{\text{emb}}}$. For a single robot $k$, the observation is ${\mathbf{S}_\textbf{MP}}_{k,t} \in   \mathbb{R}^{D_{\text{obs}} \times D_{k} }$. 

\begin{equation}
\mathbf{E}_{k,t} = \left( {\mathbf{W}_{\textbf{emb}}^\top \mathbf{S}_\textbf{MP}}_{k,t}  + \mathbf{b}_{\textbf{emb}} \right) \cdot \sqrt{{D}_{\text{emb}}}
\end{equation}

Where $\mathbf{W}_{\textbf{emb}} \in \mathbb{R}^{D_{\text{obs}} \times D_{\text{emb}}}$ and $\mathbf{b}_{\textbf{emb}} \in \mathbb{R}^{D_{\text{emb}}}$ are the weight and bias of the linear embedding layer, respectively, and they are shared parameters. While this projection maps all limb observations into a common feature dimension, the same transformation is applied to every limb. Therefore, this embedding layer does not explicitly account for limb-specific coordinate frames, joint roles, or morphology-dependent local contexts before attention-based aggregation. The multiplier $\sqrt{D_{\text{emb}}}$ scales the embedded values to counteract the increase in variance in high-dimensional space.

Each column of the embedded matrix $\mathbf{E}_{k,t} \in \mathbb{R}^{D_{\text{emb}} \times D_k}$ corresponds to a limb's feature vector. This matrix is fed into an $L^\text{th}$-layer Transformer Encoder, with the input to the first layer being $\mathbf{X}_{k,t}^{(1)}$. Since MetaMorph uses Multi-Head Self-Attention in each layer, we have $\mathbf{Q}_{k,t}^{\left ( l \right )}=\mathbf{K}_{k,t}^{\left ( l \right )}= \mathbf{V}_{k,t}^{\left ( l \right )}=\mathbf{X}_{k,t}^{\left ( l \right )}$. The encoder layers model dependencies between limbs via the self-attention mechanism. For analytical clarity, we use a linearized abstraction of the attention mixing operator and omit normalization and feed-forward nonlinearities. The resulting derivation is intended to characterize the structure of cross-token aggregation.

For layer $l\in\{1,\ldots,L\}$ and attention head
$h\in\{1,\ldots,D_h\}$, let $D_q$ and $D_v$ denote the
query/key and value dimensions, respectively. Define
\begin{align}
\mathbf{Q}_h^{(l)}
&=
{\mathbf{W}_{\mathbf Q}^{(l),h}}^\top
\mathbf{Q}_{k,t}^{(l)},\\
\mathbf{K}_h^{(l)}
&=
{\mathbf{W}_{\mathbf K}^{(l),h}}^\top
\mathbf{K}_{k,t}^{(l)},\\
\mathbf{V}_h^{(l)}
&=
{\mathbf{W}_{\mathbf V}^{(l),h}}^\top
\mathbf{V}_{k,t}^{(l)},
\end{align}
where
$\mathbf{W}_{\mathbf Q}^{(l),h},
\mathbf{W}_{\mathbf K}^{(l),h}
\in\mathbb{R}^{D_{\mathrm{emb}}\times D_q}$
and
$\mathbf{W}_{\mathbf V}^{(l),h}
\in\mathbb{R}^{D_{\mathrm{emb}}\times D_v}$.

Throughout this analysis, the first relation index denotes the source
token and the second denotes the target token. We therefore represent the
attention matrix in source-by-target orientation:
\begin{equation}
\mathbf{\Omega}_{k,t}^{(l),h}
=
\operatorname{softmax}_{\mathrm{col}}
\left(
\frac{
{\mathbf{K}_h^{(l)}}^\top
\mathbf{Q}_h^{(l)}
}{
\sqrt{D_q}
}
\right),
\label{eq:attention_source_target}
\end{equation}
where the softmax is applied independently to each target column.
The output of head $h$ is
\begin{equation}
\mathbf{H}_{k,t}^{(l),h}
=
\mathbf{V}_h^{(l)}
\mathbf{\Omega}_{k,t}^{(l),h}.
\end{equation}

Let $\alpha_{j,i}^{(l),h}$ denote the entry in row $j$ and column $i$
of $\mathbf{\Omega}_{k,t}^{(l),h}$. It represents the attention weight
from source limb $j$ to target limb $i$, and therefore satisfies
\begin{equation}
\sum_{j=1}^{D_k}\alpha_{j,i}^{(l),h}=1.
\end{equation}

Decomposing $\mathbf{H}_{k,t}$ by limb yields:

\begin{align}
\mathbf{H}_{k,t}^{\left ( l \right ),h} &= \left [ \mathbf{H}_{k,t}^{1,\left ( l \right ),h},\mathbf{H}_{k,t}^{2,\left ( l \right ),h},\cdots ,\mathbf{H}_{k,t}^{D_k,\left ( l \right ),h} \right ] \\
&= \left [ \sum_{j=1}^{D_k}\alpha _{j,1}^{\left ( l \right ),h}{\mathbf{W}_\mathbf{V}^{\left ( l \right ),h}}^\top \mathbf{V}_{k,t}^{j,\left ( l \right )} ,\sum_{j=1}^{D_k}\alpha _{j,2}^{\left ( l \right ),h}{\mathbf{W}_\mathbf{V}^{\left ( l \right ),h}}^\top\mathbf{V}_{k,t}^{j,\left ( l \right )}  ,\cdots ,\sum_{j=1}^{D_k}\alpha _{j,D_k}^{\left ( l \right ),h}{\mathbf{W}_\mathbf{V}^{\left ( l \right ),h}}^\top\mathbf{V}_{k,t}^{j,\left ( l \right )}  \right ] 
\end{align}

Concatenating the outputs of all heads along the feature dimension gives:

\begin{equation}
\mathbf{H}_{k,t}^{(l)} = \left[ {\mathbf{H}_{k,t}^{(l),1}}^\top; {\mathbf{H}_{k,t}^{(l),2}}^\top;\dots; {\mathbf{H}_{k,t}^{(l),D_h}}^\top \right]^\top \in \mathbb{R}^{\left ( D_{h}\cdot D_{v} \right ) \times D_k}.
\end{equation}

Thus, the output of the multi-head attention mechanism at layer $l$ is:

\begin{align}
\mathbf{M}_{k,t}^{(l)} &=  {\mathbf{W}_\mathbf{H}^{(l)}}^\top\mathbf{H}_{k,t}^{(l)}\\
&= {\mathbf{W}_\mathbf{H}^{(l)}}^\top\left[ {\mathbf{H}_{k,t}^{(l),1}}^\top; {\mathbf{H}_{k,t}^{(l),2}}^\top;\dots; {\mathbf{H}_{k,t}^{(l),D_h}}^\top \right]^\top  \\
&= \sum_{h=1}^{D_h}{\mathbf{W}_\mathbf{H}^{(l),h}}^\top\mathbf{H}_{k,t}^{(l),h}
\end{align}

where $\mathbf{W}_\mathbf{H}^{(l)}\in \mathbb{R}^{\left ( D_{h}\cdot D_{v} \right ) \times D_{\text{emb}}}$ is the integration weight for the multi-head attention output, which can be expanded row-wise as:

\begin{equation}
    \mathbf{W}_\mathbf{H}^{(l)}=\left [ {\mathbf{W}_\mathbf{H}^{(l),1}}^\top;{\mathbf{W}_\mathbf{H}^{(l),2}}^\top;\cdots ;{\mathbf{W}_\mathbf{H}^{(l),D_h}}^\top \right ]^\top 
\end{equation}

where $\mathbf{W}_\mathbf{H}^{(l),i}\in\mathbb{R}^{D_{v}\times D_{\textbf{emb}}}$.

The output of the multi-head attention mechanism for limb $i$ of robot $k$ at time $t$ and layer $l$ is:

\begin{align}
    \mathbf{M}_{k,t}^{i,(l)} &= \sum_{h=1}^{D_h}{\mathbf{W}_\mathbf{H}^{(l),h}}^\top\sum_{j=1}^{D_k}\alpha _{j,i}^{\left ( l \right ),h}{\mathbf{W}_\mathbf{V}^{\left ( l \right ),h}}^\top\mathbf{V}_{k,t}^{j,\left ( l \right )} \\
    &= \sum_{j=1}^{D_k}\left ( \sum_{h=1}^{D_h}\alpha _{j,i}^{(l),h}{\mathbf{W}_\mathbf{H}^{(l),h}}^\top{\mathbf{W}_\mathbf{V}^{\left ( l \right ),h}}^\top \right )  \mathbf{V}_{k,t}^{j,(l)}  \\  
    &= \sum_{j=1}^{D_k}{\mathbf{R}_{j,i}^{(l)}}^\top\mathbf{V}_{k,t}^{j,(l)}
\end{align}

Define $\mathbf{R}_{j,i}^{(l)} = \sum_{h=1}^{D_h}\alpha _{j,i}^{(l),h} \mathbf{W}_\mathbf{V}^{\left ( l \right ),h}\mathbf{W}_\mathbf{H}^{(l),h}$, $\mathbf{R}_{j,i}^{(l)}
\in \mathbb{R}^{D_{\mathrm{emb}}\times D_{\mathrm{emb}}}$ represents the multi-head aggregated effective relation matrix from limb $j$ to limb $i$ in layer $l$ of the Transformer. It quantifies the influence strength of limb $j$'s latent embedding feature $\mathbf{V}_{k,t}^{j,(l)}$ on limb $i$ during feature extraction in layer $l$.

Next, $\mathbf{M}_{k,t}^{(l)}$ is passed through a residual connection and normalization layer, then fed into a feed-forward neural network layer, followed by another residual connection and normalization layer to obtain the output of the $l^\text{th}$ Transformer encoder layer:

\begin{align}
\mathbf{Z}_{k,t}^{(l)} &= {\mathbf{W}_\mathbf{O}^{\left ( l \right )}}^\top\left ( \mathbf{M}_{k,t}^{(l)}+\mathbf{X}_{k,t}^{\left ( l \right )} \right )  + \mathbf{b}_\mathbf{O}^{\left ( l \right )} +\mathbf{M}_{k,t}^{(l)}+\mathbf{X}_{k,t}^{\left ( l \right )} \\
 &={\mathbf{W}_{\textbf{MX}}^{\left ( l \right )}}^\top \left ( \mathbf{M}_{k,t}^{(l)}+\mathbf{X}_{k,t}^{\left ( l \right )} \right )  + \mathbf{b}_\mathbf{O}^{\left ( l \right )} \\
 &= \left[{\mathbf{W}_{\textbf{MX}}^{\left ( l \right )}}^\top \left ( \sum_{j=1}^{D_k}{\mathbf{R}_{j,1}^{(l)}}^\top\mathbf{V}_{k,t}^{j,(l)}+ \mathbf{X}_{k,t}^{1,\left ( l \right )}\right ) , \cdots ,{\mathbf{W}_{\textbf{MX}}^{\left ( l \right )}}^\top \left ( \sum_{j=1}^{D_k}{\mathbf{R}_{j,D_k}^{(l)}}^\top\mathbf{V}_{k,t}^{j,(l)}+ \mathbf{X}_{k,t}^{D_k,\left ( l \right )}\right ) \right]+\mathbf{b}_\mathbf{O}^{\left ( l \right )} 
\end{align}

where $\mathbf{W}_\mathbf{O}^{\left ( l \right )}\in \mathbb{R}^{D_{\textbf{emb}}  \times D_{\textbf{emb}}}$ and $\mathbf{b}_\mathbf{O}^{\left ( l \right )} \in \mathbb{R}^{ D_{\textbf{emb}}}$ are the weight and bias of the feed-forward neural network layer, respectively. Let $\mathbf{W}_{\textbf{MX}}^{\left ( l \right )}=\mathbf{W}_\mathbf{O}^{\left ( l \right )}+\mathbf{I}$. Then the output for limb $i$ at encoder layer $l$ and time $t$ is:

\begin{align}
    \mathbf{Z}_{k,t}^{i,(l)} &= {\mathbf{W}_{\textbf{MX}}^{\left ( l \right )}}^\top \left ( \sum_{j=1}^{D_k}{\mathbf{R}_{j,i}^{(l)}}^\top\mathbf{V}_{k,t}^{j,(l)}+ \mathbf{X}_{k,t}^{i,\left ( l \right )}\right ) + \mathbf{b}_\mathbf{O}^{\left ( l \right )} \\
    &= {\mathbf{W}_{\textbf{MX}}^{\left ( l \right )}}^\top \left ( \sum_{j=1}^{D_k}{\mathbf{R}_{j,i}^{(l)}}^\top\mathbf{X}_{k,t}^{j,(l)}+ \mathbf{X}_{k,t}^{i,\left ( l \right )}\right ) + \mathbf{b}_\mathbf{O}^{\left ( l \right )} \\
    &= {\mathbf{W}_{\textbf{MX}}^{\left ( l \right )}}^\top\left ( \sum_{j=1}^{i-1}{\mathbf{R}_{j,i}^{(l)}}^\top\mathbf{X}_{k,t}^{j,(l)}+ \left ( {\mathbf{R}_{i,i}^{(l)}}^\top  +\mathbf{I} \right )\mathbf{X}_{k,t}^{i,\left ( l \right )} +\sum_{j=i+1}^{D_k}{\mathbf{R}_{j,i}^{(l)}}^\top\mathbf{X}_{k,t}^{j,(l)}\right )  + \mathbf{b}_\mathbf{O}^{\left ( l \right )} 
\end{align}

For the first encoder layer ($l=1$), $\mathbf{X}_{k,t}^{\left ( 1 \right )}=\mathbf{E}_{k,t}$. The output for limb $i$ in the first encoder layer at time $t$ is:

\begin{align}
\mathbf{Z}_{k,t}^{i,(1)}
&= {\mathbf{W}_{\textbf{MX}}^{\left ( 1 \right )}}^\top\left ( \sum_{j=1}^{i-1}{\mathbf{R}_{j,i}^{(1)}}{}^\top\mathbf{E}_{k,t}^{j}+ \left ( {\mathbf{R}_{i,i}^{(1)}}{}^\top  +\mathbf{I} \right )\mathbf{E}_{k,t}^{i} +\sum_{j=i+1}^{D_k}{\mathbf{R}_{j,i}^{(1)}}{}^\top\mathbf{E}_{k,t}^{j}\right )  + \mathbf{b}_\mathbf{O}^{\left ( 1 \right )} \\
&= {\mathbf{W}_{\textbf{MX}}^{\left ( 1 \right )}}{}^\top\left (
      \sum_{j=1}^{i-1}{\mathbf{R}_{j,i}^{(1)}}{}^\top\mathbf{W}_{\textbf{emb}}{}^\top{\mathbf{S}_\textbf{MP}}_{k,t}^{j}  
      +  \left ( {\mathbf{R}_{i,i}^{(1)}}{}^\top  +\mathbf{I} \right )\mathbf{W}_{\textbf{emb}}{}^\top{\mathbf{S}_\textbf{MP}}_{k,t}^{i}
      + \sum_{j=i+1}^{D_k}{\mathbf{R}_{j,i}^{(1)}}{}^\top\mathbf{W}_{\textbf{emb}}{}^\top{\mathbf{S}_\textbf{MP}}_{k,t}^{j}
    \right )\sqrt{D_{\text{emb}}}  \notag \\ 
&\quad
    + {\mathbf{W}_{\textbf{MX}}^{\left ( 1 \right )}}{}^\top \left (
        \sum_{j=1}^{D_k} {\mathbf{R}_{j,i}^{(1)}}{}^\top \mathbf{b}_{\textbf{emb}}
        + \mathbf{b}_{\textbf{emb}}
       \right )
       \sqrt{D_{\text{emb}}}
      + \mathbf{b}_\mathbf{O}^{(1)}
\\
&= {\mathbf{W}_{\textbf{MX}}^{\left ( 1 \right )}}{}^\top\left (
      \sum_{j=1}^{i-1}{\mathbf{\hat{R}}_{j,i}^{(1)}}{}^\top{\mathbf{S}_\textbf{MP}}_{k,t}^{j}  
      +  \left ( {\mathbf{\hat{R}}_{i,i}^{(1)}}  +\mathbf{W}_{\textbf{emb}} \right ){}^\top{\mathbf{S}_\textbf{MP}}_{k,t}^{i}
      + \sum_{j=i+1}^{D_k}{\mathbf{\hat{R}}_{j,i}^{(1)}}{}^\top{\mathbf{S}_\textbf{MP}}_{k,t}^{j}
    \right )\sqrt{D_{\text{emb}}}
\notag \\
&\quad
    + {\mathbf{W}_{\textbf{MX}}^{\left ( 1 \right )}}{}^\top\left (
        \sum_{h=1}^{D_h} {\mathbf{W}_{H}^{(1),h}}{}^\top{\mathbf{W}_\mathbf{V}^{\left ( 1 \right ),h}}{}^\top\mathbf{b}_{\textbf{emb}}
        + \mathbf{b}_{\textbf{emb}}
      \right)\sqrt{D_{\text{emb}}}
      + \mathbf{b}_\mathbf{O}^{(1)}
\\
&= {\mathbf{W}_{\textbf{MX}}^{\left ( 1 \right )}}{}^\top\left (
      \sum_{j=1}^{i-1}{\mathbf{\hat{R}}_{j,i}^{(1)}}{}^\top{\mathbf{S}_\textbf{MP}}_{k,t}^{j}  
      +  \left ( {\mathbf{\hat{R}}_{i,i}^{(1)}}  +\mathbf{W}_{\textbf{emb}} \right ){}^\top{\mathbf{S}_\textbf{MP}}_{k,t}^{i}
      + \sum_{j=i+1}^{D_k}{\mathbf{\hat{R}}_{j,i}^{(1)}}{}^\top{\mathbf{S}_\textbf{MP}}_{k,t}^{j}
    \right)\sqrt{D_{\text{emb}}}
    + \mathbf{b}_\mathbf{z}^{(1)}
\end{align}

We define

\begin{equation} 
\mathbf{\hat{R}}_{j,i}^{(l)} = \mathbf{W}_{\textbf{emb}}\mathbf{R}_{j,i}^{(l)} \end{equation}

\begin{equation} 
\mathbf{b}_\mathbf{z}^{(l)} = {\mathbf{W}_{\textbf{MX}}^{\left ( l \right )}}^\top\left (
        \sum_{h=1}^{D_h} {\mathbf{W}_{H}^{(l),h}}^\top{\mathbf{W}_\mathbf{V}^{\left ( l \right ),h}}^\top\mathbf{b}_{\textbf{emb}}
        + \mathbf{b}_{\textbf{emb}}
      \right)\sqrt{D_{\text{emb}}}
      + \mathbf{b}_\mathbf{O}^{(l)}
\end{equation}

For encoder layers $l$, $\mathbf{X}_{k,t}^{\left ( l \right )}=\mathbf{Z}_{k,t}^{\left ( l-1 \right )}$. The output for limb $i$ at encoder layer $l$ and time $t$ is:

\begin{equation}
    \mathbf{Z}_{k,t}^{i,(l)} = {\mathbf{W}_{\textbf{MX}}^{\left ( l \right )}}^\top\left ( \sum_{j=1}^{i-1}{\mathbf{R}_{j,i}^{(l)}}^\top\mathbf{Z}_{k,t}^{j,(l-1)}+ \left ( {\mathbf{R}_{i,i}^{(l)}}^\top  +\mathbf{I} \right )\mathbf{Z}_{k,t}^{i,(l-1)} +\sum_{j=i+1}^{D_k}{\mathbf{R}_{j,i}^{(l)}}^\top\mathbf{Z}_{k,t}^{j,(l-1)}\right )  + \mathbf{b}_\mathbf{O}^{\left ( l \right )}   
\end{equation}

The expansion shows that MetaMorph performs direct, state-dependent aggregation across all limb tokens. The effective relation matrix $\mathbf{R}_{j,i}^{(l)}$ varies with the current observation, providing a flexible global contextualization mechanism. RecMorph introduces a complementary structural bias by constraining feature transport to a morphology-derived order and repeatedly applying shared recurrent transitions along that path.

The final Transformer encoder layer ($L$) outputs the high-dimensional feature representation of action decisions, $\mathbf{Z}_{k,t}^{(L)}$. Since some complex tasks provide terrain information, MetaMorph performs feature encoding on the global perception information ${\mathbf{S}_\mathbf{G}}_{k,t}$ and concatenates it with $\mathbf{Z}_{k,t}^{(L)}$ along the feature dimension. The concatenated vector is then passed through an output decoder layer to obtain the predicted action distribution:

\begin{align}
\mathbf{A}_{k,t}&=\mathbf{W}_{\textbf{decoder}}^\top\left (\mathbf{Z}_{k,t}^{(L)} \oplus \left ( {\mathbf{W}_\mathbf{G}}^{\top}{\mathbf{S}_\mathbf{G}}_{k,t}+ \mathbf{b}_\mathbf{G}\right )  \right )+\mathbf{b}_{\textbf{decoder}} \\
&= {\mathbf{W}_{\textbf{decoder}}^{1}}^\top\mathbf{Z}_{k,t}^{(L)}+{\mathbf{W}_{\textbf{decoder}}^{2}}^\top\left ( {{\mathbf{W}_\mathbf{G}}^\top\mathbf{S}_\mathbf{G}}_{k,t}+ \mathbf{b}_\mathbf{G}\right )+\mathbf{b}_{\textbf{decoder}} \\
&= {\mathbf{W}_{\textbf{decoder}}^{1}}^\top\mathbf{Z}_{k,t}^{(L)}+ {\mathbf{W}_{\textbf{decoder}}^{2}}^\top{{\mathbf{W}_\mathbf{G}}^\top\mathbf{S}_\mathbf{G}}_{k,t}+ {\mathbf{W}_{\textbf{decoder}}^{2}}^\top\mathbf{b}_\mathbf{G}+\mathbf{b}_{\textbf{decoder}}\\
&= {\mathbf{W}_\mathbf{A}^{1}}^\top\mathbf{Z}_{k,t}^{(L)}+{\mathbf{W}_\mathbf{A}^{2}}^\top{\mathbf{S}_\mathbf{G}}_{k,t}+ \mathbf{b}_\mathbf{A}
\end{align}

where $\mathbf{W}_\mathbf{G}\in \mathbb{R}^{D_{G} \times D_{G}}$ and $\mathbf{b}_\mathbf{G} \in \mathbb{R}^{ D_{G}}$ are the weight and bias for encoding the global perception information, respectively. $\mathbf{W}_{\textbf{decoder}}\in \mathbb{R}^{\left ( D_{\textbf{emb}} + D_{G} \right ) \times {D_{\textbf{out}}}}$ and $\mathbf{b}_{\textbf{decoder}} \in \mathbb{R}^{D_{\textbf{out}}}$ are the weight and bias of the output decoder layer, respectively.  The symbol $\oplus$ denotes the concatenation operation along the feature dimension. $\mathbf{W}_{\textbf{decoder}}$ can be split as  $\mathbf{W}_{\textbf{decoder}} =  \left [ {\mathbf{W}_{\textbf{decoder}}^{1}}^\top;{\mathbf{W}_{\textbf{decoder}}^{2}}^\top\right ]^\top$, where $\mathbf{W}_{\textbf{decoder}}^{1} \in \mathbb{R}^{D_{\textbf{emb}} \times D_{\textbf{out}}}$,$\mathbf{W}_{\textbf{decoder}}^{2} \in \mathbb{R}^{D_{\textbf{G}} \times D_{\textbf{out}}}$. By simplifying the weights and combining biases, we define:$\mathbf{W}_\mathbf{A}^{1} = \mathbf{W}_{\textbf{decoder}}^{1}$,$\mathbf{W}_\mathbf{A}^{2} = \mathbf{W}_\mathbf{G}\mathbf{W}_{\textbf{decoder}}^{2}$,$\mathbf{b}_\mathbf{A} = {\mathbf{W}_{\textbf{decoder}}^{2}}^\top\mathbf{b}_\mathbf{G}+\mathbf{b}_{\textbf{decoder}}$.

From the expression for $\mathbf{A}_{k,t}$, it is evident that the action output by the Transformer encoder is further adjusted based on the global perception information to obtain the final action distribution.

\appsubsection{Modumorph}  
\label{subsection:Appendix_Modumorph}

This section derives the aggregation structure of ModuMorph. Compared with MetaMorph, ModuMorph introduces two morphology-conditioned components: a hypernetwork that generates limb-wise input/output projections, and an attention module whose queries and keys are conditioned on the morphological context. We analyze these components to show how ModuMorph improves the compatibility of limb-local features before aggregation.

For robot $k$ at time step $t$, in addition to the raw observation ${\mathbf{S}_\textbf{MP}}_{k,t}$, ModuMorph introduces a morphological context observation. The morphological context observation $\mathbf{C}_{k}$ extracts morphology-related information from ${\mathbf{S}_\textbf{MP}}_{k,t}^i$. For robot $k$, let
$\mathbf C_k=[\mathbf C_k^1,\ldots,\mathbf C_k^{D_k}]
\in\mathbb R^{D_{\mathrm{ctx}}\times D_k}$
denote the limb-wise morphology context. The network encodes $\mathbf{C}_{k}$ via a multi-layer MLP encoder to generate ${\mathbf{HN}_{\textbf{emb}}}_{k,t}$ and ${\mathbf{HN}_{\textbf{att}}}_{k,t}$, used for hypernetwork parameter generation and the attention mechanism, respectively:

\begin{align}
{\mathbf{HN}_{\textbf{emb}}}_{k,t} &=  \textbf{ReLU} \left(\mathbf{W}_{\textbf{emb}_\textbf{C}}^\top\mathbf{C}_{k} +\mathbf{b }_{{\textbf{emb}}_\textbf{C}}\right) \\  
{\mathbf{HN}_{\textbf{att}}}_{k,t} &=  \textbf{ReLU} \left(\mathbf{W}_{{\textbf{att}}_\textbf{C}}^\top\mathbf{C}_{k} +\mathbf{b }_{{\textbf{att}}_\textbf{C}}\right)  
\end{align}

where
$\mathbf{W}_{{\textbf{emb}}_\textbf{C}},
\mathbf{W}_{{\textbf{att}}_\textbf{C}}
\in
\mathbb{R}^{D_{\textbf{ctx}}\times
D_{\textbf{emb}_{\textbf{ctx}}}}$
and
$\mathbf{b}_{{\textbf{emb}}_\textbf{C}},
\mathbf{b}_{{\textbf{att}}_\textbf{C}}
\in
\mathbb{R}^{D_{\textbf{emb}_{\textbf{ctx}}}}$
are the weights and biases of the morphology-context encoders,
respectively.

Unlike MetaMorph's shared embedding layer, ModuMorph utilizes a hypernetwork to dynamically generate projection weights $\mathbf{W}_{\textbf{emb}_\textbf{HN}}$ and biases $\mathbf{b}_{\textbf{emb}_\textbf{HN}}$ tailored to each limb:

\begin{equation}
    {\mathbf{W}_{\textbf{emb}_\textbf{HN}}}=\mathbf{W}_{\textbf{emb}_\textbf{W}}^\top{\mathbf{HN}_{\textbf{emb}}}_{k,t}+\mathbf{b}_{\textbf{emb}_\textbf{W}}
\end{equation}

\begin{equation}
    {\mathbf{b}_{\textbf{emb}_\textbf{HN}}}=\mathbf{W}_{\textbf{emb}_\textbf{b}}^\top{\mathbf{HN}_{\textbf{emb}}}_{k,t}+\mathbf{b}_{\textbf{emb}_\textbf{b}}
\end{equation}

where $\mathbf{W}_{\textbf{emb}_\textbf{W}}\in \mathbb{R}^{D_{\textbf{emb}_{\textbf{ctx}}}\times \left ( D_{\textbf{obs}}\cdot D_{\textbf{emb}} \right )},\mathbf{W}_{\textbf{emb}_\textbf{b}}\in \mathbb{R}^{D_{\textbf{emb}_{\textbf{ctx}}}\times D_{\textbf{emb}} },\mathbf{b}_{\textbf{emb}_\textbf{W}}\in \mathbb{R}^{ \left ( D_{\textbf{obs}}\cdot D_{\textbf{emb}} \right )},\mathbf{b}_{\textbf{emb}_\textbf{b}}\in \mathbb{R}^{  D_{\textbf{emb}}  }$. $\mathbf{W}_{\textbf{emb}_\textbf{HN}}\in \mathbb{R}^{\left ( D_{\textbf{obs}}\cdot D_{\textbf{emb}} \right ) \times D_{k}}$ can be reshaped into  $\mathbf{\tilde{W} }_{\textbf{emb}_\textbf{HN}}\in \mathbb{R}^{D_{k}\times  D_{\textbf{obs}}\times D_{\textbf{emb}}}$. Using  $\mathbf{\tilde{W}}_{\textbf{emb}_\textbf{HN}}$ and $\mathbf{b}_{\textbf{emb}_\textbf{HN}}$ as the weight and bias of the linear embedding layer enables limbs of different morphological structure to possess specialized feature projection functions. Using these generated parameters, ModuMorph assigns different projection functions to different limbs according to their morphological context. This can make limb-local features more compatible before aggregation, because observations from different modules are no longer processed only by a single shared embedding matrix.

\begin{align}
 {\mathbf{E}_{\textbf{HN}}}_{k,t} = \left( \mathbf{\tilde{W}}_{\textbf{emb}_\textbf{HN}}^\top {\mathbf{S}_\textbf{MP}}_{k,t} + \mathbf{b}_{\textbf{emb}_\textbf{HN}} \right) \cdot \sqrt{D_{\text{emb}}}
\end{align}

The linear embedding output for limb $i$ of robot $k$ at time step $t$ is:

\begin{align}
 {\mathbf{E}_{\textbf{HN}}}_{k,t}^{i} = \left( {\mathbf{\tilde{W}}_{\textbf{emb}_\textbf{HN}}^{i}}{}^\top {\mathbf{S}_\textbf{MP}}_{k,t}^{i} + \mathbf{b}_{\textbf{emb}_\textbf{HN}}^{i} \right) \cdot \sqrt{D_{\text{emb}}} 
\end{align}

where $\mathbf{\tilde{W}}$ is obtained by reshaping $\mathbf{W}$, thus they are essentially the same matrix, and:

\begin{align}
\mathbf{W}_{\textbf{emb}_\textbf{HN}}^{i} 
&= \mathbf{W}_{\textbf{emb}_\textbf{W}}^\top {\mathbf{HN}_\textbf{emb}}_{ k,t}^{i} + \mathbf{b}_{\textbf{emb}_\textbf{W}}  \\ 
&= \mathbf{W}_{\textbf{emb}_\textbf{W}}^\top \text{ReLU} \left( \mathbf{W}_{\textbf{emb}_\textbf{C}}^\top \mathbf{C}_{k}^{i} + \mathbf{b}_{\textbf{emb}_\textbf{C}} \right) + \mathbf{b}_{\textbf{emb}_\textbf{W}} \\
\mathbf{b}_{\textbf{emb}_\textbf{HN}}^{i} 
&= \mathbf{W}_{\textbf{emb}_\textbf{b}}^\top {\mathbf{HN}_\textbf{emb}}_{ k,t}^{i} + \mathbf{b}_{\textbf{emb}_\textbf{b}}  \\ 
&= \mathbf{W}_{\textbf{emb}_\textbf{b}}^\top \text{ReLU} \left( \mathbf{W}_{\textbf{emb}_\textbf{C}}^\top \mathbf{C}_{k}^{i} + \mathbf{b}_{\textbf{emb}_\textbf{C}} \right) + \mathbf{b}_{\textbf{emb}_\textbf{b}}
\end{align}

Thus, ${\mathbf{E}_{\textbf{HN}}}_{k,t}^{i}$, the linear embedding for limb $i$, is generated based on its own morphological context observation $\mathbf{C}_{k}^{i}$.

ModuMorph does not use self-attention in its Transformer Encoder. Instead, it uses $\mathbf{Q}_{k,t}^{\left ( l \right )}=\mathbf{K}_{k,t}^{\left ( l \right )}={\mathbf{HN}_{\textbf{att}}}_{k,t}$, $\mathbf{V}_{k,t}^{\left ( l \right )}=\mathbf{X}_{k,t}^{\left ( l \right )}$. The output ${\mathbf{E}_{\textbf{HN}}}_{k,t}$ from the linear embedding layer serves as the input $\mathbf{X}_{k,t}^{\left ( 1 \right )}$  to the first Transformer encoder layer. For subsequent layers ($l \ge 2$), the input  $\mathbf{X}_{k,t}^{\left ( l \right )}$ is the output $\mathbf{Z}_{k,t}^{(l-1)}$ from the previous layer. Following the derivations in Appendix  \ref{subsection:Appendix_MetaMorph}, the output of head $h \in \{1, \dots, D_h\}$ in the multi-head attention mechanism for layer $l \in \{1, \dots, L\}$ is:

\begin{equation}
\mathbf{H_{\textbf{HN} }}_{k,t}^{\left ( l \right ),h} = \left ( {\mathbf{W}_\mathbf{V}^{\left ( l \right ),h}}^\top \mathbf{V}_{k,t}^{\left ( l \right )}  \right )\mathbf{\Omega_{\textbf{HN} }}_{k,t}^{\left ( l \right ),h}
\end{equation}

where 
\begin{align}
\mathbf{\Omega_{\textbf{HN} }}_{k,t}^{\left ( l \right ),h} &= \operatorname{softmax}_{\mathrm{col}}\left ( \frac{ \left ( {\mathbf{W}_\mathbf{K}^{\left ( l \right ),h}}^\top \mathbf{K}_{k,t}^{\left ( l \right )}  \right ) ^\top\left ( {\mathbf{W}_\mathbf{Q}^{\left ( l \right ),h}}^\top \mathbf{Q}_{k,t}^{\left ( l \right )}  \right )}{\sqrt{D_q} }  \right ) \\
&= \operatorname{softmax}_{\mathrm{col}}\left ( \frac{\left ( {\mathbf{W}_\mathbf{K}^{\left ( l \right ),h}}^\top {\mathbf{HN_{\textbf{att}}}}_{k,t}  \right )^\top \left ( {\mathbf{W}_\mathbf{Q}^{\left ( l \right ),h}}^\top {\mathbf{HN_{\textbf{att}}}}_{k,t}  \right ) }{\sqrt{D_q} }  \right )  
\end{align}

Let
${\alpha_{\mathrm{HN}}}_{j,i}^{(l),h}$
denote the entry in row $j$ and column $i$ of
$\mathbf{\Omega}_{\mathrm{HN},k,t}^{(l),h}$.
It represents the morphology-conditioned attention weight from
source limb $j$ to target limb $i$. Then, the output for limb $i$ in the first Transformer encoder layer at time $t$ is:

\begin{align}
{\mathbf{Z}_{\textbf{HN} }}_{k,t}^{i,(1)}= {\mathbf{W}_{\textbf{MX}}^{(1)}}^\top \left ( \sum_{j=1}^{i-1}{\mathbf{\hat{R}}_{\textbf{HN},j,i}^{(1)}}{}^\top {\mathbf{S}_\textbf{MP}}_{k,t}^{j}
      + \left ( {\mathbf{\hat{R}}_{\textbf{HN},i,i}^{(1)}}{} + \mathbf{\tilde{W}}_{\textbf{emb}_\textbf{HN}}^{i} \right )^\top {\mathbf{S}_\textbf{MP}}_{k,t}^{i}
      + \sum_{j=i+1}^{D_k}{\mathbf{\hat{R}}_{\textbf{HN},j,i}^{(1)}}{}^\top {\mathbf{S}_\textbf{MP}}_{k,t}^{j}
     \right )  \sqrt{D_{\text{emb}}}
    + \mathbf{b_{\textbf{HN}_\textbf{z}}}^{i,(1)} 
\end{align}

where
\begin{align} 
\mathbf{\hat{R}}_{\textbf{HN},j,i}^{(l)} &= \mathbf{\tilde{W}}_{\textbf{emb}_\textbf{HN}}^{j}\mathbf{R}_{\textbf{HN},j,i}^{(l)} \\
 &= \mathbf{\tilde{W}}_{\textbf{emb}_\textbf{HN}}^{j}\sum_{h=1}^{D_h}{\alpha_{\text{HN}}}_{j,i}^{(l),h} \mathbf{W}_\mathbf{V}^{\left ( l \right ),h}\mathbf{W}_{H}^{(l),h} 
\end{align}

\begin{equation} 
\mathbf{b}_\mathbf{\textbf{HN}_\textbf{z}}^{i,(l)} = {\mathbf{W}_{\textbf{MX}}^{\left ( l \right )}}^\top \left ( \sum_{h=1}^{D_h}\sum_{j=1}^{D_k} {\alpha_{\text{HN}}}_{j,i}^{(l),h} {\mathbf{W}_\mathbf{H}^{(l),h}}^\top {\mathbf{W}_\mathbf{V}^{\left ( l \right ),h}}^\top \mathbf{b}_\mathbf{\textbf{emb}_\textbf{HN}}^{j} + \mathbf{b}_\mathbf{\textbf{emb}_\textbf{HN}}^{i}\right )  \sqrt{\mathbf{D}_{\textbf{emb} } } +\mathbf{b}_\mathbf{O}^{\left ( l \right )} 
\end{equation}

For encoder layers $l \ge 2$, where $\mathbf{X}_{k,t}^{\left ( l \right )}=\mathbf{Z}_{k,t}^{\left ( l-1 \right )}$, the output for limb $i$ in layer $l$ at time $t$ is:

\begin{align}
    {\mathbf{Z}_{\textbf{HN} }}_{k,t}^{i,(l)} &= {\mathbf{W_{\textbf{MX}}}^{\left ( l \right )}}^\top \left ( \sum_{j=1}^{i-1}{\mathbf{R}_{\textbf{HN},j,i}^{(l)}}^\top \mathbf{Z}_{k,t}^{j,(l-1)}+ \left ( {\mathbf{R_{\textbf{HN}}}_{i,i}^{(l)}} +\mathbf{I} \right )^\top \mathbf{Z}_{k,t}^{i,(l-1)} +\sum_{j=i+1}^{D_k}{\mathbf{R}_{\textbf{HN},j,i}^{(l)}}^\top \mathbf{Z}_{k,t}^{j,(l-1)}\right )  + \mathbf{b}_\textbf{O}^{\left ( l \right )}  
\end{align}

Because $\mathbf{R}_{\mathrm{HN},j,i}^{(l)}$ is conditioned on morphology, ModuMorph provides a morphology-dependent aggregation pattern that remains fixed for a given body. Together with its generated limb-wise projections, this mechanism improves cross-limb representation compatibility before and during global aggregation. RecMorph uses morphology differently: kinematic structure determines the path along which shared recurrent transitions transport and contextualize limb features.

Similarly, the output decoder layer in ModuMorph dynamically generates projection weights $\mathbf{W}_{\textbf{decoder}_\textbf{HN}}$ and biases $\mathbf{b}_{\textbf{decoder}_\textbf{HN}}$ for each limb via a hypernetwork:

\begin{equation}
     {\mathbf{W}_{\textbf{decoder}_\textbf{HN}}}=\mathbf{W}_{\textbf{decoder}_\textbf{W}}^\top {\mathbf{HN}_{\textbf{emb}}}_{k,t}+\mathbf{b}_{\textbf{decoder}_\textbf{W}} 
\end{equation}
\begin{equation}
     {\mathbf{b}_{\textbf{decoder}_\textbf{HN}}}=\mathbf{W}_{\textbf{decoder}_\textbf{b}}^\top {\mathbf{HN}_{\textbf{emb}}}_{k,t}+\mathbf{b}_{\textbf{decoder}_\textbf{b}}
\end{equation}

where $\mathbf{W}_{\textbf{decoder}_\textbf{W}}\in \mathbb{R}^{D_{\textbf{emb}_{\textbf{ctx}}}\times \left ( \left (  D_{\textbf{emb}} + D_{G} \right )  \cdot D_{\textbf{out}} \right )},\mathbf{W}_{\textbf{decoder}_\textbf{b}}\in \mathbb{R}^{D_{\textbf{emb}_{\textbf{ctx}}}\times   D_{\textbf{out}} },\mathbf{b}_{\textbf{decoder}_\textbf{W}}\in \mathbb{R}^{\left (  D_{\textbf{emb}} + D_{G} \right )  \cdot D_{\textbf{out}}},\mathbf{b}_{\textbf{decoder}_\textbf{b}}\in \mathbb{R}^{  D_{\textbf{out}}  }$.$\mathbf{W}_{\mathrm{decoderHN}}
\in
\mathbb{R}^{
((D_{\mathrm{emb}}+D_G)D_{\mathrm{out}})
\times D_k
}$.
Each column is reshaped into a limb-specific decoder matrix
$\tilde{\mathbf W}_{\mathrm{decoderHN}}^{i}
\in
\mathbb{R}^{(D_{\mathrm{emb}}+D_G)\times D_{\mathrm{out}}}$. The generated parameters are reshaped into limb-specific decoder weights and biases, providing each limb with a morphology-conditioned action projection from the shared contextual representation. For tasks providing global perception, ModuMorph encodes the global perception information ${\mathbf{S}_\mathbf{G}}_{k,t}$, concatenates it with ${\mathbf{Z}_{\textbf{HN} }}_{k,t}^{(L)}$ along the feature dimension, and finally passes the concatenated vector through the output decoder layer to obtain the predicted action distribution:

\begin{align}
\mathbf{A_\textbf{HN}}_{k,t}&={\mathbf{W}_{\textbf{decoder}_\textbf{HN}}}^\top \left ({\mathbf{Z}_{\textbf{HN} }}_{k,t}^{(L)} \oplus \left ( {\mathbf{W}_\mathbf{G}}^\top {\mathbf{S}_\mathbf{G}}_{k,t}+ \mathbf{b}_\mathbf{G}\right )  \right )+\mathbf{b}_{\textbf{decoder}_\textbf{HN}} \\
&= {\mathbf{W}_{\textbf{decoder}_\textbf{HN}}^{1}}^\top {\mathbf{Z}_{\textbf{HN} }}_{k,t}^{(L)} + {\mathbf{W}_{\textbf{decoder}_\textbf{HN}}^{2}}^\top \left ( {\mathbf{W}_\mathbf{G}}^\top {\mathbf{S}_\mathbf{G}}_{k,t}+ \mathbf{b}_\mathbf{G}\right ) +\mathbf{b}_{\textbf{decoder}_\textbf{HN}} \\
&= {\mathbf{W}_{\textbf{decoder}_\textbf{HN}}^{1}}^\top {\mathbf{Z}_{\textbf{HN} }}_{k,t}^{(L)} + {\mathbf{W}_{\textbf{decoder}_\textbf{HN}}^{2}}^\top {\mathbf{W}_\mathbf{G}}^\top {\mathbf{S}_\mathbf{G}}_{k,t}+ {\mathbf{W}_{\textbf{decoder}_\textbf{HN}}^{2}}^\top \mathbf{b}_\mathbf{G}+\mathbf{b}_{\textbf{decoder}_\textbf{HN}}\\
&= {\mathbf{W}_{A_\textbf{HN}}^{1}}^\top {\mathbf{Z}_{\textbf{HN} }}_{k,t}^{(L)} + {\mathbf{W}_{A_\textbf{HN}}^{2}}^\top {\mathbf{S}_\mathbf{G}}_{k,t} + \mathbf{b}_{A_\textbf{HN}}
\end{align}

\appsubsection{NerveNet}
\label{subsection:Appendix_NerveNet}

We analyze a linearized abstraction of the NerveNet-style message-passing operator to contrast graph-local aggregation with RecMorph's ordered recurrent transport.

For robot $k$ at time step $t$, let $\mathbf{h}_{k,t}^{i,(l)} \in \mathbb{R}^{D_{\textbf{node}}}$ denote the hidden state feature vector of limb $i$ at the $l^\text{th}$ message-passing iteration. The initial state $\mathbf{h}_{k,t}^{i,(0)}$ is derived from the linear embedding of the raw observation ${\mathbf{S}_\textbf{MP}}_{k,t}^i$. 

In NerveNet, the graph propagation relies on the adjacency matrix $\mathbf{a} \in \mathbb{R}^{D_k \times D_k}$ defined by the robot's morphology, where $a_{i,j}$ indicates the connection strength from limb $j$ to limb $i$. The update rule for node $i$ involves message generation, aggregation, self-feature transformation, and a residual update. To isolate the neighborhood aggregation structure, we analyze a linearized
form of the message-passing update and omit nonlinear activations and
normalization from the algebraic expansion.

First, each node projects its current feature into a message space. For a neighboring limb $j$, the generated message is:
\begin{equation}
    \mathbf{m}_{k,t}^{j,(l)} = {\mathbf{W}_{\textbf{proj}}}^\top \mathbf{h}_{k,t}^{j,(l)} + \mathbf{b}_{\textbf{proj}}
\end{equation}
where $\mathbf{W}_{\textbf{proj}} \in \mathbb{R}^{D_{\textbf{node}} \times D_{\textbf{msg}}}$ and $\mathbf{b}_{\textbf{proj}} \in \mathbb{R}^{D_{\textbf{msg}}}$ are shared projection parameters.

The messages from all neighboring nodes $j \in \mathcal{N}(i)$ are then aggregated via isotropic weighted summation based on the adjacency matrix:
\begin{equation}
    \bar{\mathbf{m}}_{k,t}^{i,(l)} = \sum_{j \in \mathcal{N}(i)} a_{i,j} \mathbf{m}_{k,t}^{j,(l)} = \sum_{j \in \mathcal{N}(i)} a_{i,j} \left( {\mathbf{W}_{\textbf{proj}}}^\top \mathbf{h}_{k,t}^{j,(l)} + \mathbf{b}_{\textbf{proj}} \right)
\end{equation}

Simultaneously, node $i$ performs a linear transformation on its own feature representation:
\begin{equation}
    \mathbf{s}_{k,t}^{i,(l)} = {\mathbf{W}_{\textbf{self}}}^\top \mathbf{h}_{k,t}^{i,(l)} + \mathbf{b}_{\textbf{self}}
\end{equation}

The self-feature $\mathbf{s}_{k,t}^{i,(l)}$ and the aggregated neighbor message $\bar{\mathbf{m}}_{k,t}^{i,(l)}$ are concatenated and passed through a shared message-fusion layer. Let $\mathbf{W}_{\textbf{msg}} \in \mathbb{R}^{(D_{\textbf{node}} + D_{\textbf{msg}}) \times D_{\textbf{node}}}$ be the weight matrix of this fusion layer. We can split $\mathbf{W}_{\textbf{msg}}$ along the feature dimension as $\mathbf{W}_{\textbf{msg}} = \left[ {\mathbf{W}_{\textbf{msg\_self}}}^\top ; {\mathbf{W}_{\textbf{msg\_neigh}}}^\top \right]^\top$. The fused output $\mathbf{z}_{k,t}^{i,(l)}$ is:
\begin{align}
    \mathbf{z}_{k,t}^{i,(l)} &= {\mathbf{W}_{\textbf{msg}}}^\top \left( \mathbf{s}_{k,t}^{i,(l)} \oplus \bar{\mathbf{m}}_{k,t}^{i,(l)} \right) + \mathbf{b}_{\textbf{msg}} \\
    &= {\mathbf{W}_{\textbf{msg\_self}}}^\top \mathbf{s}_{k,t}^{i,(l)} + {\mathbf{W}_{\textbf{msg\_neigh}}}^\top \bar{\mathbf{m}}_{k,t}^{i,(l)} + \mathbf{b}_{\textbf{msg}}
\end{align}

Finally, NerveNet utilizes a residual connection via an output projection layer to update the node representation for the next iteration:
\begin{equation}
    \mathbf{h}_{k,t}^{i,(l+1)} = \mathbf{h}_{k,t}^{i,(l)} + {\mathbf{W}_{\textbf{out}}}^\top \mathbf{z}_{k,t}^{i,(l)} + \mathbf{b}_{\textbf{out}}
\end{equation}

By substituting the expanded forms of $\mathbf{s}_{k,t}^{i,(l)}$ and $\bar{\mathbf{m}}_{k,t}^{i,(l)}$ into the final update equation, we obtain the holistic expression:
\begin{align}
    \mathbf{h}_{k,t}^{i,(l+1)} 
    &= \mathbf{h}_{k,t}^{i,(l)} + {\mathbf{W}_{\textbf{out}}}^\top \left( {\mathbf{W}_{\textbf{msg\_self}}}^\top \left( {\mathbf{W}_{\textbf{self}}}^\top \mathbf{h}_{k,t}^{i,(l)} + \mathbf{b}_{\textbf{self}} \right) \right. \notag \\
    &\quad \left. + {\mathbf{W}_{\textbf{msg\_neigh}}}^\top \left( \sum_{j \in \mathcal{N}(i)} a_{i,j} \left( {\mathbf{W}_{\textbf{proj}}}^\top \mathbf{h}_{k,t}^{j,(l)} + \mathbf{b}_{\textbf{proj}} \right) \right) + \mathbf{b}_{\textbf{msg}} \right) + \mathbf{b}_{\textbf{out}} \\
    &= \left( \mathbf{I} + {\mathbf{W}_{\textbf{out}}}^\top {\mathbf{W}_{\textbf{msg\_self}}}^\top {\mathbf{W}_{\textbf{self}}}^\top \right) \mathbf{h}_{k,t}^{i,(l)} \notag \\
    &\quad + \left( {\mathbf{W}_{\textbf{out}}}^\top {\mathbf{W}_{\textbf{msg\_neigh}}}^\top {\mathbf{W}_{\textbf{proj}}}^\top \right) \sum_{j \in \mathcal{N}(i)} a_{i,j} \mathbf{h}_{k,t}^{j,(l)} + \mathbf{B}_{\textbf{total}}^{(l)}
\end{align}

To reveal the algebraic essence of the GNN propagation, we define the composite transformation matrices:
\begin{equation}
    \mathbf{W}_{\textbf{ST}} = \mathbf{W}_{\textbf{self}} \mathbf{W}_{\textbf{msg\_self}} \mathbf{W}_{\textbf{out}}
\end{equation}
\begin{equation}
    \mathbf{W}_{\textbf{MT}} = \mathbf{W}_{\textbf{proj}} \mathbf{W}_{\textbf{msg\_neigh}} \mathbf{W}_{\textbf{out}}
\end{equation}
and the consolidated bias term:
\begin{align}
    \mathbf{B}_{\textbf{total}}^{(l)} &= {\mathbf{W}_{\textbf{out}}}^\top {\mathbf{W}_{\textbf{msg\_self}}}^\top \mathbf{b}_{\textbf{self}} \notag \\
    &\quad + {\mathbf{W}_{\textbf{out}}}^\top {\mathbf{W}_{\textbf{msg\_neigh}}}^\top \left( \sum_{j \in \mathcal{N}(i)} a_{i,j} \mathbf{b}_{\textbf{proj}} \right) + {\mathbf{W}_{\textbf{out}}}^\top \mathbf{b}_{\textbf{msg}} + \mathbf{b}_{\textbf{out}}
\end{align}

This simplifies the NerveNet update mechanism into the following mathematically transparent form:
\begin{equation}
    \mathbf{h}_{k,t}^{i,(l+1)} = \left( \mathbf{I} + {\mathbf{W}_{\textbf{ST}}}^\top \right) \mathbf{h}_{k,t}^{i,(l)} + {\mathbf{W}_{\textbf{MT}}}^\top \sum_{j \in \mathcal{N}(i)} a_{i,j} \mathbf{h}_{k,t}^{j,(l)} + \mathbf{B}_{\textbf{total}}^{(l)}
\end{equation}

\textbf{Interpretation of the Message-Passing Operator.}
The derived update shows that NerveNet aggregates neighboring information through the weighted summation
$\sum_{j \in \mathcal{N}(i)} a_{i,j}\mathbf{h}_{k,t}^{j,(l)}$
followed by a shared projection $\mathbf{W}_{\mathbf{MT}}$. Increasing graph-message-passing depth expands the receptive field, but
each additional layer aggregates representations that already contain
neighborhood mixtures from previous layers. Consequently, global
communication and repeated neighborhood mixing grow together with message
passing depth. RecMorph also performs repeated transformations, but the repetition occurs
along an ordered hidden-state trajectory inside one spatial sequence
operator instead of through successively deeper neighborhood-pooling
layers.

\appsection{Linearized Analysis of Topology-Guided Recurrent Transport}
\label{app:linearized_transport}

Under the linearized analysis, RMS normalization and the SiLU input
nonlinearity are absorbed into an effective input-to-hidden map. Relative
to Sec.~\ref{sec:spatial_recurrence}, the recurrent parameters correspond to
\begin{align}
{\overrightarrow{\mathbf W}_{h}^{(l)}}^\top
&\equiv A_{\rightarrow}^{(l)},&
{\overrightarrow{\mathbf W}_{x}^{(l)}}^\top
&\equiv B_{\rightarrow}^{(l)}W_x^{(l)},\\
{\overleftarrow{\mathbf W}_{h}^{(l)}}^\top
&\equiv A_{\leftarrow}^{(l)},&
{\overleftarrow{\mathbf W}_{x}^{(l)}}^\top
&\equiv B_{\leftarrow}^{(l)}W_x^{(l)}.
\end{align}

Let $\mathbf{X}_{k,t}^{i,(l)}$ denote the representation presented
to the recurrent transition at token $i$ and block $l$.
The output projection
${{\mathbf W}_{\mathrm{out}}^{(l)}}^\top$
corresponds to the linearized bidirectional contextual projection
before residual integration. Bias terms are omitted from the
source-dependent influence expansion because they contribute additive
terms independent of the source-token index.

The policy network first projects the raw observation ${\mathbf{S}_\textbf{MP}}_{k,t}$ for each limb of robot $k$ at time step $t$ onto a latent space using a linear embedding layer to obtain the embedded vector $\mathbf{E}_{k,t}$. The embedded limb vectors $\mathbf{E}_{k,t}$ are then fed into the BiRNN. Unlike the global, parallel attention of the Transformer, the BiRNN models inter-limb dependencies via bidirectional recursive information propagation along the morphological topological order. We use Depth-First Search (DFS) by default to form the morphological topological order, as DFS naturally places physically connected limbs consecutively in the sequence, preserving the locality of limb combinations. This ordering allows the recurrent backbone to propagate information through sequence neighborhoods that often correspond to physically connected or nearby modules. We denote the input to the $l^\text{th}$ layer as $\mathbf{X}_{k,t}^{(l)}$, where $\mathbf{X}_{k,t}^{(1)}=\mathbf{E}_{k,t}$.  

The BiRNN layer consists of two independent directions. The forward
recurrence processes the sequence from $1$ toward $D_k$, while the
backward recurrence processes it from $D_k$ toward $1$. Consequently,
token $i$ incorporates context from tokens $1,\ldots,i$ through the
forward state and from tokens $i,\ldots,D_k$ through the backward state.

\begin{equation}
\overrightarrow{\mathbf{h}}_{k,t}^{i,(l)} = \sigma\left( {\overrightarrow{\mathbf{W}}_\mathbf{h}^{(l)}}^\top \overrightarrow{\mathbf{h}}_{k,t}^{i-1,(l)} + {\overrightarrow{\mathbf{W}}_\mathbf{x}^{(l)}}^\top \mathbf{X}_{k,t}^{i,(l)} + \overrightarrow{\mathbf{b}}^{(l)}\right)   
\end{equation}
\begin{equation}
\overleftarrow{\mathbf{h}}_{k,t}^{i,(l)} = \sigma\left( {\overleftarrow{\mathbf{W}}_{h}^{(l)}}^\top \overleftarrow{\mathbf{h}}_{k,t}^{i+1,(l)} + {\overleftarrow{\mathbf{W}}_{x}^{(l)}}^\top \mathbf{X}_{k,t}^{i,(l)} + \overleftarrow{\mathbf{b}}^{(l)}\right) 
\end{equation}

Here, $\overrightarrow{\mathbf{h}}_{k,t}^{i,(l)}$ and $\overleftarrow{\mathbf{h}}_{k,t}^{i,(l)}$ represent the hidden states for limb $i$ in the forward and backward processes, respectively. $\overrightarrow{\mathbf{W}}_\mathbf{x}^{(l)}$ and $\overleftarrow{\mathbf{W}}_\mathbf{x}^{(l)}$ are the projection matrices from input to hidden state for the forward and backward processes, respectively. $\overrightarrow{\mathbf{W}}_\mathbf{h}^{(l)}$ and $\overleftarrow{\mathbf{W}}_\mathbf{h}^{(l)}$ are the recurrent weight matrices between hidden states for the forward and backward processes, respectively. $\overrightarrow{\mathbf{b}}^{(l)}$ and $\overleftarrow{\mathbf{b}}^{(l)}$ are the bias vectors for the forward and backward processes, respectively. $\sigma$ is a nonlinear activation function. For derivation coherence, we assume the initial states $\overrightarrow{\mathbf{h}}_{k,t}^{0,(l)}$ and $\overleftarrow{\mathbf{h}}_{k,t}^{D_k+1,(l)}$ are zero vectors and approximate $\sigma$ as a linear function.

At layer $l$, the output $\mathbf{Z}_{k,t}^{i,(l)}$ for limb $i$ is obtained by concatenating the forward and backward hidden states and applying a linear projection:

\begin{equation}
\mathbf{Z}_{k,t}^{i,(l)} = {\mathbf{W}_{\mathbf{out}}^{(l)}}^\top \left( \overrightarrow{\mathbf{h}}_{k,t}^{i,(l)} \oplus \overleftarrow{\mathbf{h}}_{k,t}^{i,(l)} \right) + \mathbf{b}_{\mathbf{out}}^{(l)}
\end{equation}

where $\oplus$ denotes vector concatenation, $\mathbf{W}_{\mathbf{out}}^{(l)}$ is the output projection matrix, and $\mathbf{b}_{\mathbf{out}}^{(l)}$ is the output projection bias. At this point, $\mathbf{Z}_{k,t}^{i,(l)}$ already incorporates information from the entire morphological sequence. To characterize the sequence-distance-dependent transport induced by the BiRNN, we recursively expand the hidden states.

Taking the forward process as an example, the hidden state $\overrightarrow{\mathbf{h}}_{k,t}^{i,(l)}$ for limb $i$ can be expanded as a cumulative function of inputs from preceding limbs. Approximating by ignoring the nonlinear effects of the activation function for linear analysis, we have:

\begin{align}
\overrightarrow{\mathbf{h}}_{k,t}^{i,(l)} &\approx \sum_{j=1}^{i} {\left( {\overrightarrow{\mathbf{W}}_{\mathbf{h}}^{(l)}}^\top \right)}^{i-j} {\overrightarrow{\mathbf{W}}_{\mathbf{x}}^{(l)}}^\top \mathbf{X}_{k,t}^{j,(l)} \\
&= {\overrightarrow{\mathbf{W}}_{\mathbf{x}}^{(l)}}^\top \mathbf{X}_{k,t}^{i,(l)} + {\overrightarrow{\mathbf{W}}_{\mathbf{h}}^{(l)}}^\top {\overrightarrow{\mathbf{W}}_{\mathbf{x}}^{(l)}}^\top \mathbf{X}_{k,t}^{i-1,(l)} + \cdots + {\left( {\overrightarrow{\mathbf{W}}_{\mathbf{h}}^{(l)}}^\top \right)}^{i-1} {\overrightarrow{\mathbf{W}}_{\mathbf{x}}^{(l)}}^\top \mathbf{X}_{k,t}^{1,(l)}
\end{align}

Similarly, the backward hidden state expands as:

\begin{align}
\overleftarrow{\mathbf{h}}_{k,t}^{i,(l)} &\approx \sum_{j=i}^{D_k} {\left( {\overleftarrow{\mathbf{W}}_{\mathbf{h}}^{(l)}}^\top \right)}^{j-i} {\overleftarrow{\mathbf{W}}_{\mathbf{x}}^{(l)}}^\top \mathbf{X}_{k,t}^{j,(l)}
\end{align}

The output feature $\mathbf{Z}_{k,t}^{i,(l)}$ for limb $i$ of robot $k$ at time $t$ in the $l^\text{th}$ BiRNN layer takes the form

\begin{align}
\mathbf{Z}_{k,t}^{i,(l)} &= {\mathbf{W}_{\mathbf{out}}^{(l)}}^\top \left( \overrightarrow{\mathbf{h}}_{k,t}^{i,(l)} \oplus \overleftarrow{\mathbf{h}}_{k,t}^{i,(l)} \right) + \mathbf{b}_{\mathbf{out}}^{(l)}  \\
&= {\overrightarrow{\mathbf{W}}_{\mathbf{out}}^{(l)}}^\top\sum_{j=1}^{i} {\left( {\overrightarrow{\mathbf{W}}_{\mathbf{h}}^{(l)}}^\top \right)}^{i-j} {\overrightarrow{\mathbf{W}}_{\mathbf{x}}^{(l)}}^\top \mathbf{X}_{k,t}^{j,(l)}+{\overleftarrow{\mathbf{W}}_{\mathbf{out}}^{(l)}}^\top \sum_{j=i}^{D_k} {\left( {\overleftarrow{\mathbf{W}}_{\mathbf{h}}^{(l)}}^\top \right)}^{j-i} {\overleftarrow{\mathbf{W}}_{\mathbf{x}}^{(l)}}^\top \mathbf{X}_{k,t}^{j,(l)} + \mathbf{b}_{\mathbf{out}}^{(l)}  \\
&= \sum_{j=1}^{i-1}{\overrightarrow{\mathbf{W}}_{\mathbf{out}}^{(l)}}^\top{\left( {\overrightarrow{\mathbf{W}}_{\mathbf{h}}^{(l)}}^\top \right)}^{i-j} {\overrightarrow{\mathbf{W}}_{\mathbf{x}}^{(l)}}^\top \mathbf{X}_{k,t}^{j,(l)}+\left ( {\overrightarrow{\mathbf{W}}_{\mathbf{out}}^{(l)}}^\top{\overrightarrow{\mathbf{W}}_{\mathbf{x}}^{(l)}}^\top+ {\overleftarrow{\mathbf{W}}_{\mathbf{out}}^{(l)}}^\top{\overleftarrow{\mathbf{W}}_{\mathbf{x}}^{(l)}}^\top\right )  \mathbf{X}_{k,t}^{i,(l)} \notag  \\
& \quad + \sum_{j=i+1}^{D_k} {\overleftarrow{\mathbf{W}}_{\mathbf{out}}^{(l)}}^\top{\left( {\overleftarrow{\mathbf{W}}_{\mathbf{h}}^{(l)}}^\top \right)}^{j-i} {\overleftarrow{\mathbf{W}}_{\mathbf{x}}^{(l)}}^\top \mathbf{X}_{k,t}^{j,(l)} + \mathbf{b}_{\mathbf{out}}^{(l)}
\end{align}

Here, $\left( {\mathbf{W}_{\mathbf{h}}^{(l)}}^\top \right)^{i-j}$ and $\left( {\mathbf{W}_{\mathbf{h}}^{(l)}}^\top \right)^{j-i}$ are defined as equivalent recurrent propagation weight matrices. For $j<i$, the source-to-target contribution is weighted by
\[
{\overrightarrow{\mathbf W}_{\mathrm{out}}^{(l)}}^\top
\left(
{\overrightarrow{\mathbf W}_{h}^{(l)}}^\top
\right)^{i-j}
{\overrightarrow{\mathbf W}_{x}^{(l)}}^\top,
\]
with the analogous backward expression for $j>i$. Thus, the number of repeated recurrent transformations is determined by the source-target separation in the serialized morphology. This gives the serialized
morphology a direct role in shaping the depth of feature transport between
limbs.  This analysis characterizes how repeated shared transitions progressively contextualize limb-local representations before action decoding. Unlike input-dependent global attention, the recurrent transition parameters are shared along the morphology-derived sequence, providing a consistent structural bias across bodies.

The high-dimensional feature representation $\mathbf{Z}_{k,t}^{(L)}$ output by the $L^\text{th}$ BiRNN encoder layer serves as the preliminary action decision feature. Similar to the Transformer version, when global terrain information is available, RecMorph encodes the global perception information ${\mathbf{S}_\mathbf{G}}_{k,t}$ and concatenates it with $\mathbf{Z}_{k,t}^{(L)}$, finally passing it through a decoder layer to obtain the action distribution prediction:

\begin{align}
\mathbf{A}_{k,t}&=\mathbf{W}_{\textbf{decoder}}^\top \left (\mathbf{Z}_{k,t}^{(L)} \oplus \left ( {\mathbf{W}_\mathbf{G}}^\top {\mathbf{S}_\mathbf{G}}_{k,t}+ \mathbf{b}_\mathbf{G}\right ) \right ) +\mathbf{b}_{\textbf{decoder}} \\
&= {\mathbf{W}_{\textbf{decoder}}^{1}}^\top \mathbf{Z}_{k,t}^{(L)} + {\mathbf{W}_{\textbf{decoder}}^{2}}^\top \left ( {\mathbf{W}_\mathbf{G}}^\top {\mathbf{S}_\mathbf{G}}_{k,t}+ \mathbf{b}_\mathbf{G}\right ) +\mathbf{b}_{\textbf{decoder}} \\
&= {\mathbf{W}_\mathbf{A}^{1}}^\top \mathbf{Z}_{k,t}^{(L)} + {\mathbf{W}_\mathbf{A}^{2}}^\top {\mathbf{S}_\mathbf{G}}_{k,t}+ \mathbf{b}_\mathbf{A}
\end{align}

where $\mathbf{W}_\mathbf{G}$ are the global perception encoding parameters, and $\mathbf{W}_{\textbf{decoder}}$ are the decoder layer parameters. The decoder matrix is partitioned into $\mathbf{W}_{\textbf{decoder}}^{1}$ for processing proprioceptive features and $\mathbf{W}_{\textbf{decoder}}^{2}$ for encoded environmental features.

In summary, topology-guided bidirectional recurrence provides an ordered
feature-transport operator whose effective transformation depth varies
with the relative position of limb tokens in the morphology-derived
sequence. The resulting contextual representation is subsequently combined
with task-specific exteroceptive information, and decoded
into the shared action space.

\appsection{Implementation Details in the UNIMAL space}  
\label{subsection:Appendix_Implementation_Details}

The UNIMAL experiments use a common PPO optimization protocol across
methods, with benchmark architecture settings matched to prior
generalized morphology controllers where applicable. In this experiment, within each task, all morphologies are optimized through the same shared
actor--critic using the PPO configuration in
Table~\ref{tab:ppo_hyperparams_unimal}; no additional morphology-specific
optimization procedure is introduced. Table~\ref{tab:ppo_hyperparams_unimal} summarizes the shared PPO
configuration. For methods using KL-based early stopping, the threshold
$\delta$ is selected from $\{0.03,0.05\}$ using the same task-specific
search space; the selected values are reported in
Table~\ref{tab:early_stopping_threshold}. RecMorph-specific architectural
settings are listed in Table~\ref{tab:RecMorph_hyperparams_unimal}.

\begin{table}[h]
\centering
\begin{tabular}{lr}
\toprule
\textbf{Hyperparameter Name} & \textbf{Hyperparameter Value} \\
\midrule
Num of Random Seeds & 4 \\
Discount $\gamma$ & 0.99 \\
GAE Parameter $\lambda$ & 0.95 \\
Policy Epochs & 8 \\
Batch Size & 5120 \\
Number of Parallel Environments & 32 \\
Total Timesteps & $1 \times 10^8$ \\
Optimizer & Adam \\
Initial Learning Rate & 0.0003 \\
Learning Rate Schedule & Linear warmup and cosine decay \\
Warmup Iterations & 5 \\
Gradient Clipping ($l_2$ norm) & 0.5 \\
Value Loss Coefficient & 0.2 \\
\bottomrule
\end{tabular}
\caption{Hyperparameter settings for PPO in UNIMAL.}
\label{tab:ppo_hyperparams_unimal}
\end{table}

\begin{table}[h]
\centering
\begin{tabular}{crrrr}
\toprule
\textbf{Environment} & \textbf{MetaMorph*} & \textbf{SWAT} & \textbf{ModuMorph} & \textbf{RecMorph} \\
\midrule
FT & 0.05 & 0.05 & 0.05 & 0.05 \\
INCLINE & 0.03 & 0.03 & 0.05 & 0.05 \\
VT & 0.03 & 0.03 & 0.03 & 0.03 \\
OBSTACLES & 0.03 & 0.03 & 0.03 & 0.03 \\
EXPLORATION & 0.03 & 0.03 & 0.03 & 0.03 \\
\bottomrule
\end{tabular}
\caption{Optimal value of the early stopping threshold for each method.}
\label{tab:early_stopping_threshold}
\end{table}

\begin{table}[h]
\centering
\begin{tabular}{lr}
\toprule
\textbf{Hyperparameter Name} & \textbf{Hyperparameter Value} \\
\midrule
Linear Projector layers & 1 \\
Linear hidden dim & 128 \\
Backbone layers & 4 \\
Sequence model hidden dim & 256 \\
Normalization & RMSNorm \\
Feature modulation & SiLU channel modulation \\
Sequence model activation & tanh \\
Decoder layers & 1 \\
\bottomrule
\end{tabular}
\caption{Hyperparameter settings for RecMorph in UNIMAL.}
\label{tab:RecMorph_hyperparams_unimal}
\end{table}

\appsection{Details of morphological information removal in UNIMAL design space}
\label{subsection:Appendix_Details_of_morphological_information_removal_in_UNIMAL_design_space}

We partition the morphology observation into topology-dependent geometric
attributes and topology-independent hardware descriptors. The ablation
removes the selected position- and orientation-related fields listed in
Table~\ref{tab:topology_removal_features} while retaining mass, body shape,
joint range, joint axis, and gear parameters. Architecture-level structural
information, including morphology-derived token ordering, is unchanged.

\begin{table}[t]
\begin{tabular}{ccc}
\hline
\textbf{Feature Category}                              & \textbf{Full Morphology} & \textbf{Attribute-Ablated Morphology Observation } \\ \hline
{Limb Model (Topology-Dependent)}       & body\_pos                              & {-}                                                  \\
                                                       & body\_ipos                             &                                                                     \\
                                                       & body\_iquat                            &                                                                     \\
                                                       & geom\_quat                             &                                                                     \\ \hline
{Limb Hardware (Topology-Independent)}  & body\_mass                             & body\_mass                                                          \\
                                                       & body\_shape                            & body\_shape                                                         \\ \hline
Joint Model (Topology-Dependent)                       & jnt\_pos                               & -                                                                   \\ \hline
{Joint Hardware (Topology-Independent)} & joint\_range                           & joint\_range                                                        \\
                                                       & joint\_axis                            & joint\_axis                                                         \\
                                                       & gear                                   & gear                                                                \\ \hline
\end{tabular}
\centering
\caption{Morphology attributes retained under the topology-attribute ablation.}
\label{tab:topology_removal_features}
\footnotesize{Note: Topology-dependent features reflect relative positional and orientational relationships between limbs, while Topology-independent features only describe inherent hardware properties.}
\end{table}

\appsection{Feature-Space Analysis of Cross-Limb Contextualization}
\label{app:feature_space}

\begin{figure}[h]
  \centering
  \includegraphics[width=0.8\columnwidth]{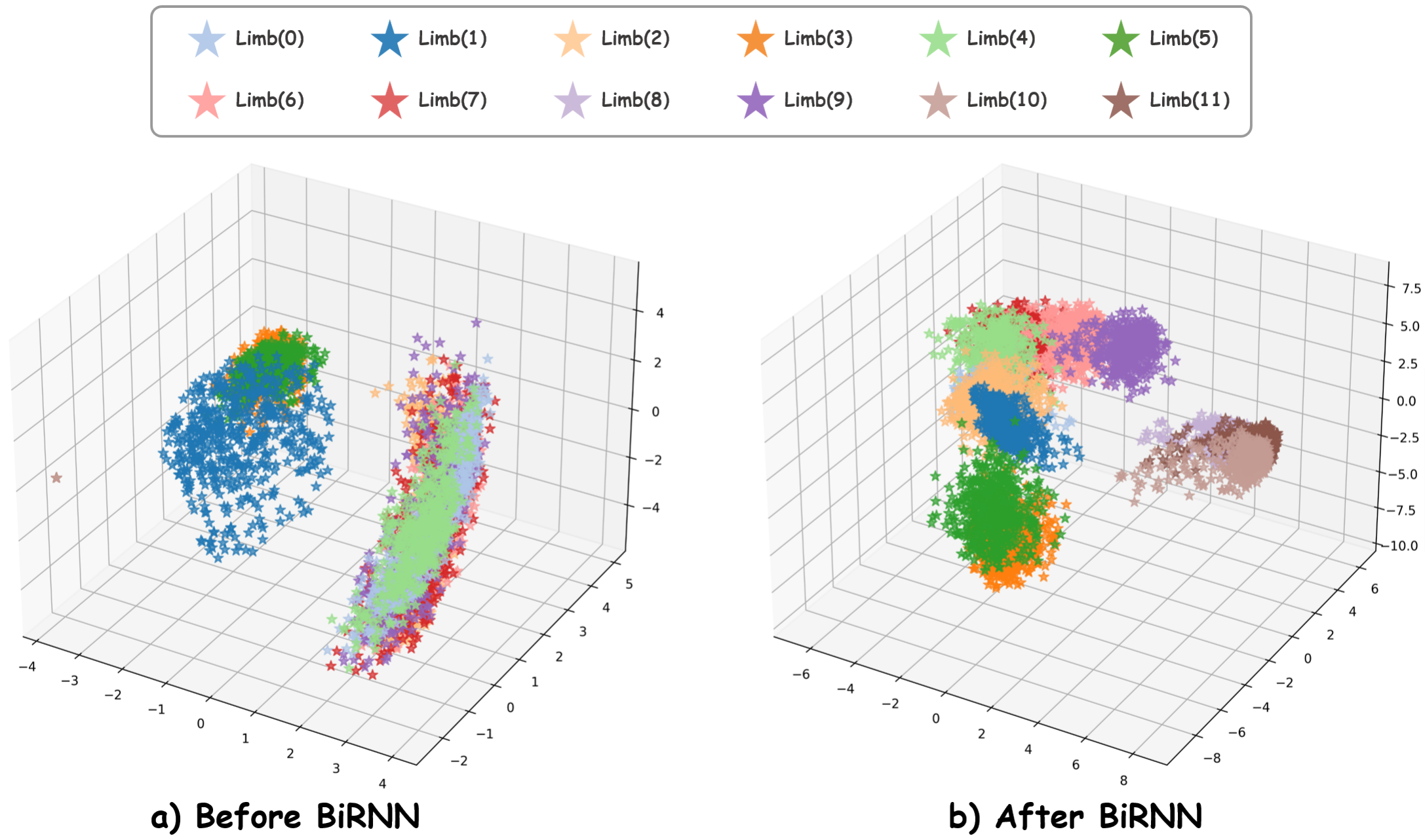}
  \caption{
  Feature-space visualization before and after recurrent transport. Limb-identity clusters become less separated after communication, consistent with increased cross-limb contextualization.
  }
  \label{fig:visual_pca}
\end{figure}

We analyze how recursive propagation changes the structure of limb features. We sample robots from the training distribution and run each instance for 500 steps, using the first 200 steps for pre-equilibration and the remaining 300 steps for feature collection. We extract features from two stages: pre-recursion features before the BiRNN and routed features during recursive propagation.

We use the Calinski--Harabasz (CH) index~\cite{de2026improving} to quantify the separability of limb-identity clusters, complemented by PCA visualizations. A higher CH value indicates that features are more separable by limb identity, while a lower value indicates that features from different limbs become more mixed in the latent space. The CH index decreases from 3747.81 before recurrence to 1436.39 after recurrent transport, indicating reduced separability of limb-identity clusters. PCA shows the same qualitative trend. These observations complement the frozen-encoder probes in Sec.~\ref{subsec:representation_alignment}: communication reduces explicit identity structure while the main-text probes show that the resulting features become more predictive of the policy actions.

\appsection{Sensitivity of RecMorph to Topology-Dependent Morphology Attributes}

\label{subsection:Appendix_The_influence_of_morphological_topology_in_the_UNIMAL_space_on_RecMorph_in_different_tasks}    

\begin{figure*}[h]  
  \vspace{-0.1in}    
  \centering
  \includegraphics[width=\linewidth]{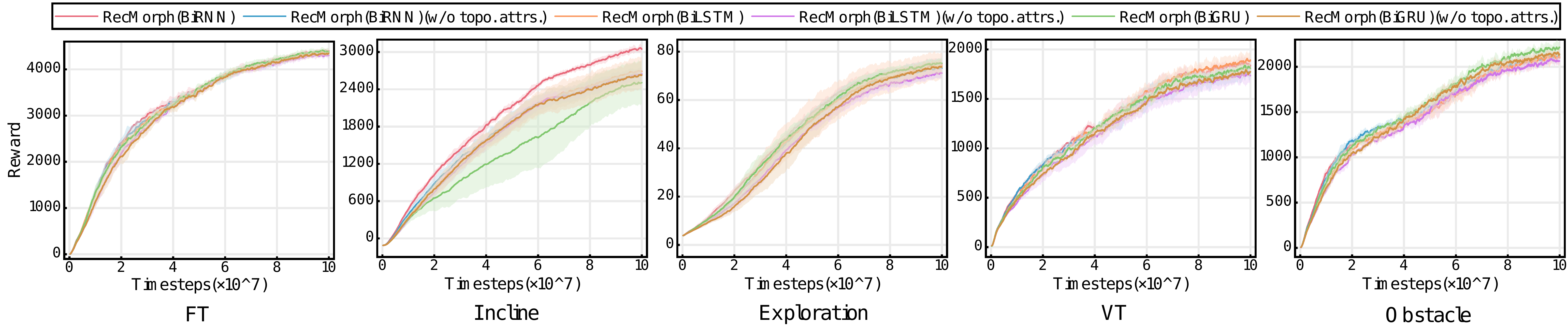} 
  \caption{The performance of RecMorph on five terrains after removing topology-dependent morphology attributes from the UNIMAL space.} 
  \label{fig:section_A_E}  
  \vspace{-0.2in}    
\end{figure*}

We next apply the same attribute-removal intervention to RecMorph while
retaining its DFS-derived communication order. Figure~\ref{fig:section_A_E}
shows task-dependent sensitivity to the selected topology-dependent
morphology attributes. FT changes little, whereas larger differences appear
on the remaining tasks, with the strongest observed effect on Incline.
These results show that RecMorph draws morphological information from two
complementary sources: explicit morphology attributes embedded in each
token and topology-derived structure encoded by the recurrent communication
order.

\appsection{Backbone Extensibility with Bidirectional Mamba2}
\label{subsection:Appendix_The_underlying_sequence_model_of_RecMorph_is_Mamba2}  

\begin{figure}[h]
    \centering
    \includegraphics[width=\textwidth]{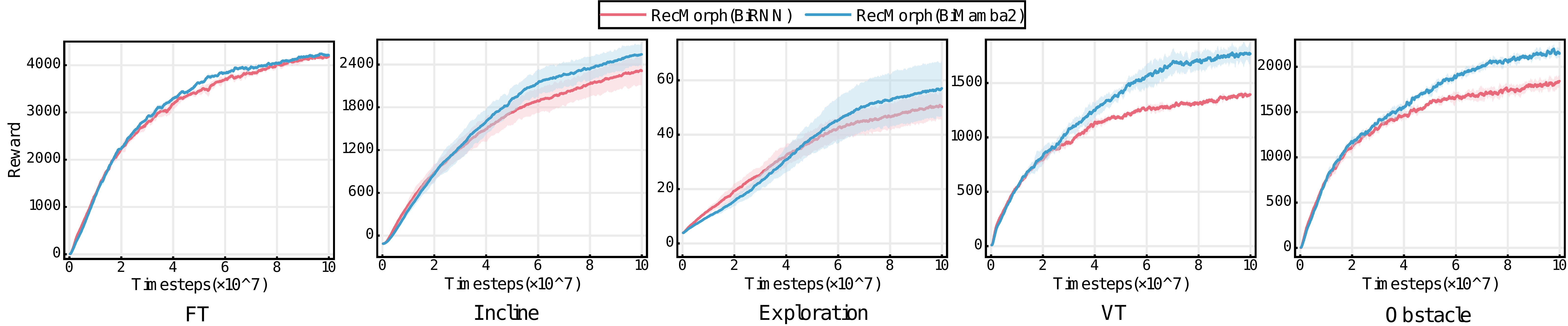}
    \caption{\textbf{Comparison between BiRNN and BiMamba2 as the underlying recursive model of RecMorph.} All variants are configured with 2 layers for a fair comparison under memory constraints.}
    \label{fig:BiRNN_and_MAMBA2}
\end{figure}

We further evaluate the backbone extensibility of RecMorph by replacing
the BiRNN transition with a bidirectional Mamba2 operator. Both BiRNN and
BiMamba2 are configured with two sequence layers to match the computational
budget used in this comparison. As shown in
Fig.~\ref{fig:BiRNN_and_MAMBA2}, BiMamba2 achieves higher training returns
across the five tasks, with particularly clear improvements on Incline and
Obstacle. These results show that morphology-guided spatial transport is
not tied to a specific recurrent cell and can accommodate alternative
sequence operators.

\appsection{Comparison with Graph-Based Message Passing}
\label{app:graph_message_passing}

The linearized NerveNet operator in
Appendix~\ref{subsection:Appendix_NerveNet} aggregates neighboring
representations through a permutation-invariant neighborhood operator
followed by shared feature transformations. Repeated message passing
therefore expands the receptive field over the kinematic graph while
preserving graph-local communication. Each additional message-passing layer aggregates representations that
already contain neighborhood mixtures from earlier layers. Receptive-field
growth and repeated local mixing therefore increase together with
message-passing depth. RecMorph also applies repeated transformations, but these transformations
occur along an ordered hidden-state trajectory within a spatial sequence
operator rather than through successively deeper neighborhood pooling. The comparison with NerveNet in
Fig.~\ref{fig:average_rewards_training_ICRA} shows that this ordered
recurrent communication mechanism achieves stronger control performance
on the evaluated UNIMAL tasks.

\appsection{Scalability to Larger Morphology Graphs}
\label{app:larger_morphologies}  

\begin{figure}[h]
    \vspace{-0.2in}
    \centering
    \includegraphics[width=0.5\columnwidth]{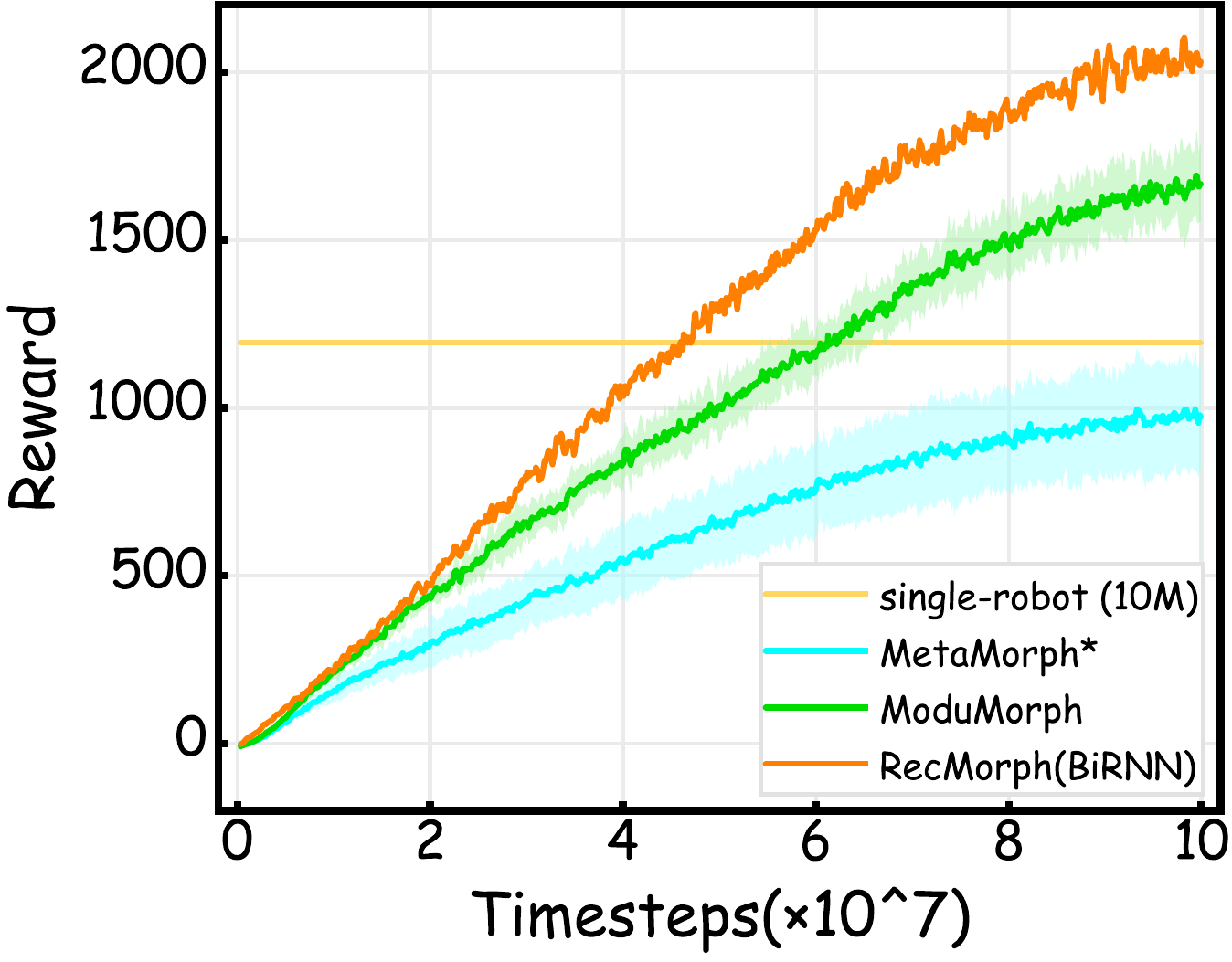}
    \caption{Training results on the newly generated multi-limb dataset on FT terrains.}  
    \label{fig:new_datasets_ft}
    \vspace{-0.2in}
\end{figure}

To further stress-test scalability, we construct a larger UNIMAL-style dataset with up to 30 limbs and an average of 25 limbs, compared with the original dataset whose maximum and average limb counts are 12 and 10, respectively. This setting substantially increases the sequence length and control complexity relative to the original benchmark. Figure~\ref{fig:new_datasets_ft} shows that all methods achieve lower returns on the larger-limb dataset than on the original benchmark. This degradation should not be attributed solely to long-range information attenuation, because even the single-robot (10M) MLP baseline, which does not rely on sequential message passing, drops from 4671 on the original dataset to 1194 on the larger-limb dataset. This indicates that increasing the number of limbs also increases the intrinsic difficulty of the control problem. Despite the increased control difficulty, RecMorph retains the strongest
learning curve among the evaluated shared controllers while also achieving
the highest measured throughput. On the larger-body dataset, MetaMorph*,
ModuMorph, and RecMorph(BiRNN) operate at 1224, 918, and 1428 FPS,
respectively. The result is consistent with the linear token-complexity
motivation of recurrent spatial transport as morphology size increases.

\appsection{Robustness to Partial Observation Loss}
\label{app:sensor_dropout}  

\begin{figure*}[h]  
  \vspace{-0.1in}    
  \centering
  \includegraphics[width=0.5\linewidth]{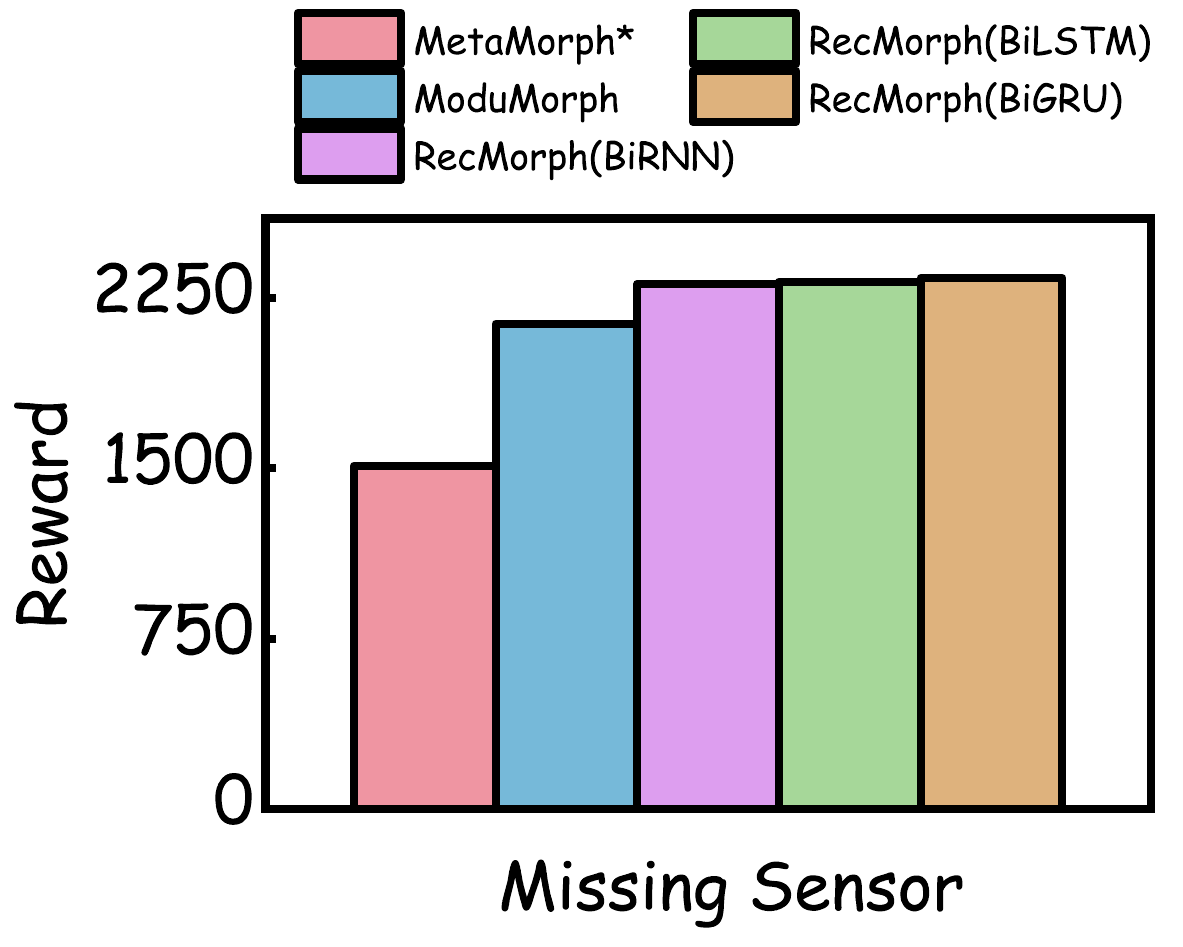} 
  \caption{FT performance under single-limb observation dropout}
  \label{fig:Zero_Shot_non_sensor} 
\end{figure*}

To evaluate robustness under realistic sensor failures, we simulate partial observation loss without altering the physical structure of the robot. Specifically, for each robot, we randomly select one limb and set its observation input to zero during execution, while keeping the physical body intact in the environment. This setup reflects real-world scenarios where sensors may malfunction while the limb continues to affect dynamics through mass, inertia, and collisions.  Figure~\ref{fig:Zero_Shot_non_sensor} shows that RecMorph retains stronger performance than the evaluated baselines under single-limb observation dropout. This behavior is consistent with body-wide contextualization allowing information from other limbs to compensate for missing local observations.

\appsection{Extension to Closed-Loop Kinematic Structures}
\label{sec:closed-loop-kinematic-structures}  

\begin{figure*}[h]  
  \vspace{-0.1in}    
  \centering
  \includegraphics[width=0.5\linewidth]{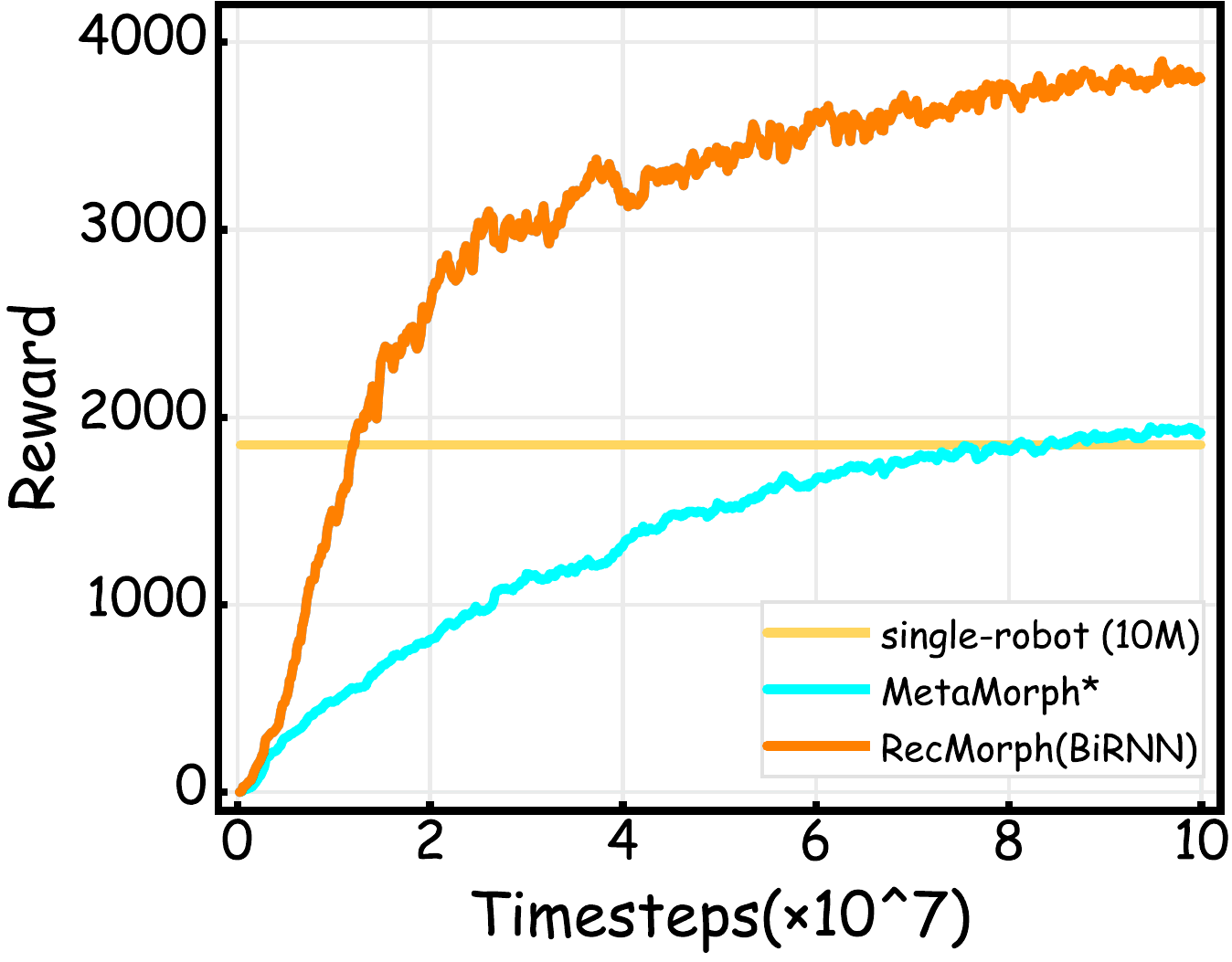} 
  \caption{Experiments on a generic controller conducted on an FT landscape using a dataset containing closed-loop data.} 
  \label{fig:loop_train}  
  \vspace{-0.2in}    
\end{figure*}

We extend RecMorph to closed-loop kinematic graphs through spanning-tree
serialization. The benchmark contains 10 newly generated closed-loop
morphologies and 40 tree-structured UNIMAL morphologies. For each
closed-loop graph, a spanning tree is constructed first and DFS is then
applied to obtain the recurrent communication order. RecMorph and the
comparison controllers are trained under the same protocol.
Figure~\ref{fig:loop_train} shows that RecMorph retains effective
locomotion performance on this mixed benchmark and outperforms the
evaluated baselines. The result demonstrates that spanning-tree
serialization provides a practical extension of the RecMorph computation
path to the evaluated closed-loop morphologies.

\appsection{Qualitative Locomotion Visualization}
\label{app:gait}  

\begin{table}[t]
\centering
\caption{
Hardware deployment statistics. Each trial lasts 15~s. Success denotes completing the trial without falling or triggering emergency stop.
}
\label{tab:hardware_results}
\begin{tabular}{lccc}
\toprule
Robot & Setting & Trials & Success \\
\midrule
Go1 & backward-right walking & 10 & 10/10 \\
Go2 & forward walking & 10 & 10/10 \\
Go2 & masked feet & 10 & 10/10 \\
Go2 & plastic-wrapped foot & 10 & 10/10 \\
\bottomrule
\end{tabular}
\end{table}

\begin{figure*}[h]  
  \vspace{-0.1in}    
  \centering
  \includegraphics[width=\linewidth]{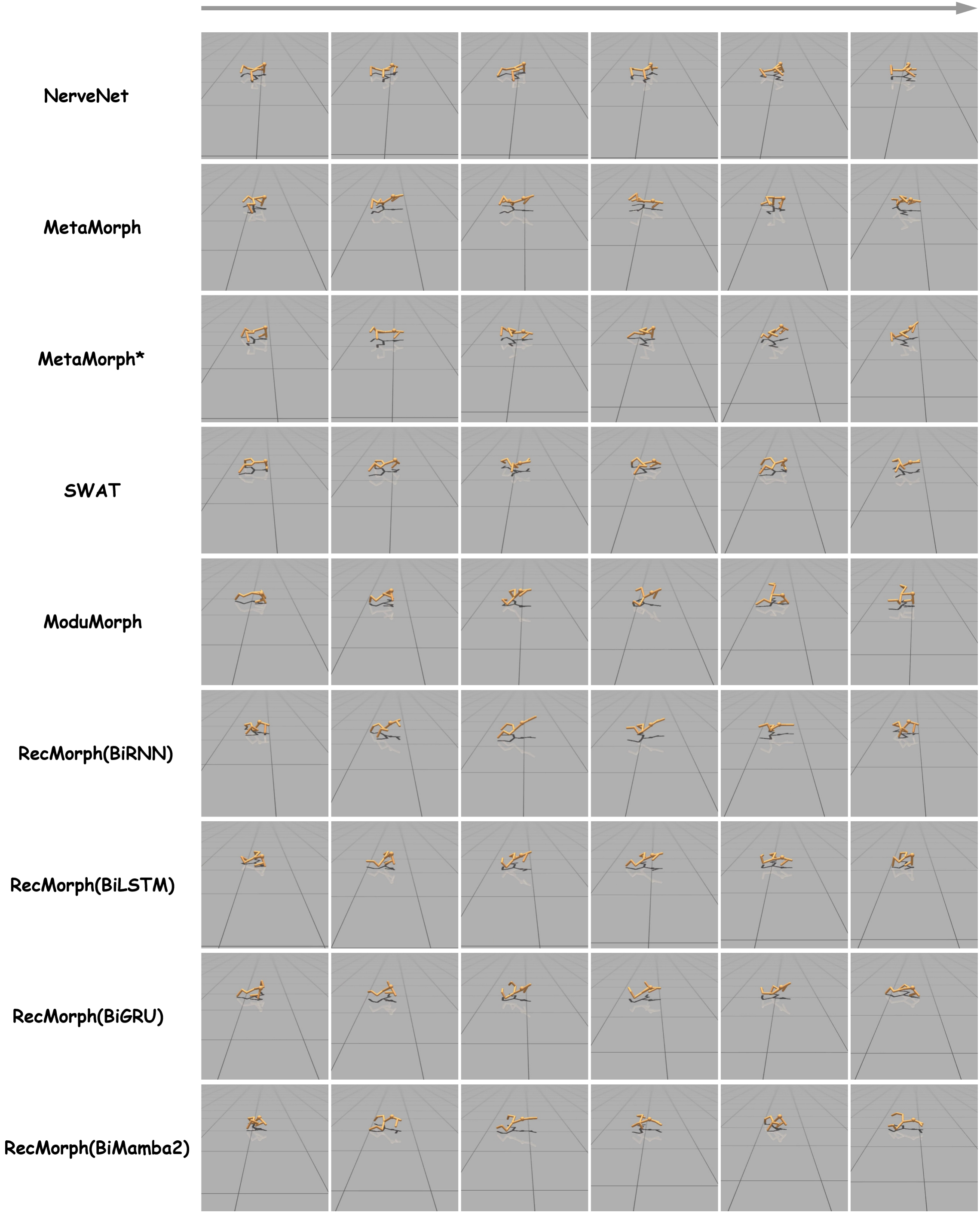} 
  \caption{Dynamic gait graphs for RecMorph and baseline methods. These graphs are temporally continuous, proceeding sequentially from left to right.} 
  \label{fig:act_gait}  
  \vspace{-0.2in}    
\end{figure*}

We provide dynamic gait visualizations (Figure \ref{fig:act_gait}) comparing RecMorph with baseline methods. The visualizations qualitatively show more regular and coordinated
locomotion patterns for RecMorph in the illustrated episodes. These
examples are intended as qualitative complements to the quantitative
benchmark results.

\appsection{Quadruped Training and Deployment Details}
\label{appendix:quadruped_details}

\paragraph{Task and platforms.}
We evaluate shared quadruped control across Unitree Go1, Unitree Go2,
ANYmal-B, and ANYmal-C. Each platform has 12 actuated joints and is
trained for velocity-tracking locomotion through a common token-based
policy interface. Robot-specific joint indexing, nominal configurations,
low-level gains, and safety constraints remain in the actuation layer.

\paragraph{Cross-platform training protocol.}
All evaluated methods are trained with 128 environments per robot,
32 rollout steps, and 10,000 PPO iterations. Four independent random
seeds are used for RecMorph and every comparison controller. Shared-policy
methods jointly optimize one controller across all four platforms, whereas
the specialist MLP baseline trains an independent controller for each
robot.

\paragraph{Friction evaluation.}
Cross-platform robustness is evaluated under low, nominal, and high
static/dynamic friction coefficients of $(0.3,0.2)$, $(0.8,0.6)$, and
$(1.2,1.0)$, respectively. The metrics reported in
Tables~\ref{tab:friction} and~\ref{tab:friction_all} are velocity-tracking
RMSE, forward progress, body-tilt RMS, and fall rate, with all aggregate
statistics computed across the four robot platforms.

\paragraph{Observation and action interface.}
All platforms use a fixed token-based interface with 64 tokens and
32 features per token. Twelve tokens are active for each quadruped and
the remaining tokens are padding. Padded dimensions are masked from action
outputs, entropy computation, and value aggregation. Active tokens contain
proprioceptive and command information available to the physical
controller, including base motion, projected gravity, velocity commands,
joint states, previous actions, joint-limit features, and the active mask.

\paragraph{Physical deployment.}
Physical Go1 and Go2 experiments use a separately trained shared RecMorph
checkpoint with policy seed 1409. The policy produces joint-position
offsets according to
\begin{equation}
q_{\mathrm{target},k}
=
q_{\mathrm{default},k}
+
0.25\,a_{\theta,k}.
\end{equation}
The controller runs at 50~Hz. Training includes flat and micro-rough
terrain together with corrupted proprioceptive observations to support
physical deployment. Robot-specific nominal poses, joint indexing,
low-level gains, and safety limits remain unchanged.

\appsection{Hardware Deployment Statistics}
\label{appendix:hardware_results}

We report trial-level deployment statistics for the physical robot experiments in Table~\ref{tab:hardware_results}. Each trial lasts 15~s. A trial is considered successful if the robot completes the commanded motion without falling or triggering emergency stop. Across all four evaluated hardware settings, the deployed RecMorph policy completes all 40 trials successfully.

\endgroup

\clearpage

\end{document}